\documentclass[english]{article}

\pdfoutput=1

\newcommand{\swaptext}[2]{#2} 

\swaptext{
\usepackage[]{neurips_2020}
}
{
\usepackage[preprint]{neurips_2020}
}

\usepackage{graphicx}
\usepackage{amssymb}
\usepackage{amsmath,amssymb,amsthm}
\usepackage{slashed}
\swaptext{ % black links for NeurIPS
\usepackage[colorlinks=true,linkcolor=black,anchorcolor=black,citecolor=black,filecolor=black,menucolor=black,runcolor=black,urlcolor=black]{hyperref}
}
{ % blue links for NeurIPS
\usepackage[colorlinks=true,urlcolor=blue,anchorcolor=blue,citecolor=blue,filecolor=blue,linkcolor=blue,menucolor=blue,linktocpage=true]{hyperref}
}
\usepackage[T1]{fontenc}
\usepackage{caption}
\usepackage{subcaption}
\usepackage{verbatim}
\usepackage{cite}

\newcommand{\cN}{\mathcal{N}}
\newcommand{\cO}{\mathcal{O}}
\newcommand{\cP}{\mathcal{P}}

\newcommand{\bR}{\mathbb{R}}

\usepackage[dvipsnames]{xcolor}
\definecolor{darkgreen}{HTML}{008000}

\newcommand{\dho}{\partial}

\newcommand{\lexpp}[1]{\mathbb{E}_{#1}\left[}
\newcommand{\rexp}{\right]}

\newtheorem{theorem}{Theorem}
\newtheorem{conj}{Conjecture}

\newtheorem{defn}{Definition}
\newtheorem{lemma}{Lemma}
\newtheorem{cor}{Corollary}

\usepackage{booktabs} % for professional tables

\title{On the asymptotics of wide networks with polynomial activations}

\author{
  Kyle Aitken \thanks{Work done while an intern at Google.}\\
  Department of Physics \\
  University of Washington \\
  Seattle, WA\\
  \texttt{kaitken17@gmail.com} \\
  \And
  Guy Gur-Ari \\
  Google\\
  Mountain View, CA\\
  \texttt{guyga@google.com} \\
}

\begin{document}

\maketitle

\begin{abstract}
We consider an existing conjecture addressing the asymptotic behavior of neural networks in the large width limit.
The results that follow from this conjecture include tight bounds on the behavior of wide networks during stochastic gradient descent, and a derivation of their finite-width dynamics.
We prove the conjecture for deep networks with polynomial activation functions, greatly extending the validity of these results.
Finally, we point out a difference in the asymptotic behavior of networks with analytic (and non-linear) activation functions and those with piecewise-linear activations such as ReLU.
\end{abstract}

% \tableofcontents{}

%%%%%%%%%%%%%%%%%%%%%%%%%%%%%%%%%%%%%%%%%%%%%%%%
\section{Introduction}

The wide network limit is interesting for both practical and theoretical reasons.
On the practical side, there is significant empirical evidence that neural networks tend to have improved performance as their width becomes large \citep{neyshabur2017exploring,zagoruyko2016wide,belkin2018reconciling}.
On the theoretical side, at large width one can gain analytic control over network dynamics, both at initialization and during training \citep{lee2017deep,ntk}.
In particular, infinitely wide networks trained using gradient flow behave as linear models with random features \citep{2019arXiv190206720L}.

\citet{dyer_asymptotics_2019} presented a mathematical tool that allows one to derive the asymptotic behavior of a large class of functions called \emph{correlation functions}.
These functions involve the network function $f_\theta(x)$ and its derivatives with respect to the network parameters $\theta$.
Here are a few examples of correlation functions, which are functions of several network inputs.
\begin{subequations}
\label{eq:ex_correlation_fns}
\begin{align}
  C_{2,0}(x_1,x_2) &:= \lexpp{\theta} f_\theta(x_1) f_\theta(x_2) \rexp \,, \label{eq:C20} \\
  C_{2,1}(x_1,x_2) &:= \sum_\mu \lexpp{\theta}
  \dho_\mu f_\theta(x_1) \dho_\mu f_\theta(x_2) \rexp \,, \label{eq:NTK} \\
  C_{4,2}(x_1,x_2,x_3,x_4) &:= \sum_{\mu_1,\mu_2} \lexpp{\theta}
  \dho_{\mu_1} \dho_{\mu_2} f_\theta(x_1) \dho_{\mu_1} f_\theta(x_2) \dho_{\mu_2} f_\theta(x_3) f_\theta(x_4)
  \rexp \,. \label{eq:C42}
\end{align}
\end{subequations}
In the above, $\dho_\mu f(x) := \dho f(x) / \dho \theta^\mu$, the index $\mu$ runs over the set of all network parameters, and $\lexpp{\theta} \cdot \rexp$ denotes the mean over initializations that we assume are Gaussian and i.i.d. 
Such correlation functions often arise in the study of the dynamics of the network function under stochastic gradient descent (SGD).
For example, \eqref{eq:NTK} is the Neural Tangent Kernel (NTK) \citep{ntk}, which controls the dynamics of gradient flow at infinite width.

The main result of \citet{dyer_asymptotics_2019} was a conjecture that argued correlation functions obey an asymptotic bound $\cO(n^s)$ where $n$ is the width of the network $f_\theta$ and $s$ is an easily computable exponent.
They showed that the conjecture can be used to prove that infinitely wide networks behave as linear models when trained with SGD (extending the gradient flow results of \citet{ntk}).
In addition, it was shown that one can use the conjecture to derive the optimization trajectory of networks with finite width, leading to a better analytic understanding of the dynamics of networks with practical width.
We believe that these results establish this conjecture as an important tool in the theoretical study of deep networks.
Therefore, proving the conjecture for more general cases is of interest, and is the purpose of this work.

\begin{figure}
\centering
\begin{subfigure}[b]{0.3\textwidth}
    \centering
    \includegraphics[scale=0.5]{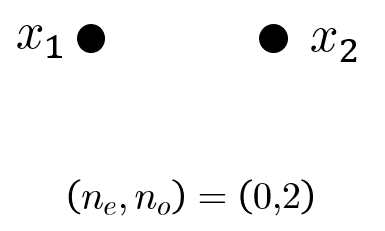}
    \caption{}
\end{subfigure}
\hfill
\begin{subfigure}[b]{0.3\textwidth}
    \centering
    \includegraphics[scale=0.5]{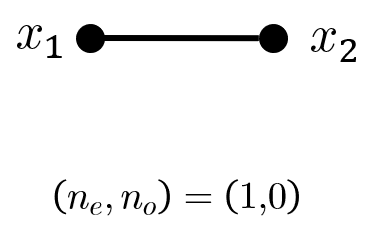}
    \caption{}
\end{subfigure}
\hfill
\begin{subfigure}[b]{0.3\textwidth}
    \centering
    \includegraphics[scale=0.5]{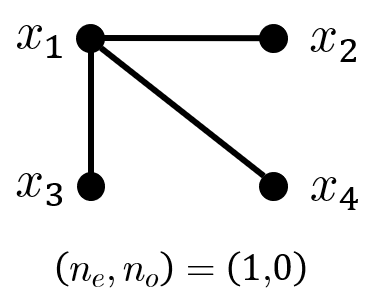}
    \caption{}
\end{subfigure}
\hfill
\caption{The cluster graphs and associated $(n_e,n_o)$ of the correlation functions in \eqref{eq:ex_correlation_fns}.
Using the conjecture, the asymptotic behavior of these correlation functions are $\cO(n^0)$, $\cO(n^0)$, and $\cO(n^{-1})$, respectively.
\label{fig:example cluster graphs}}
\end{figure}

\paragraph*{Review of the conjecture.}
Suppose $f_\theta(x)$ is the scalar output of a neural network, where $x\in \mathbb{R}^d$ is the input and $\theta$ are the model parameters.
In this work we consider the asymptotic behavior of \textit{correlation functions}, a class of functions involving $f_\theta$ and its derivatives with respect to the model parameters, in the large width limit.
A general correlation function $C$ takes the schematic form
\begin{align}
\label{eq:gen_C}
  C(x_1,\dots,x_m) = \sum_{\rm indices} \lexpp{\theta} \dho^{k_1} f_\theta(x_1) \cdots \dho^{k_m} f_\theta(x_m) \rexp \,.
\end{align}
Here $x_1,\dots,x_m$ are network inputs, and we use $\partial^k f_\theta(x)$ as shorthand for the rank-$k$ derivative tensor $\dho^{k} f_\theta(x)/\partial\theta^{\mu_1}\cdots\partial\theta^{\mu_{k}}$.
In particular, $\dho^0 f_\theta=f_\theta$ is the network function itself, $\dho^1 f_\theta$ is the gradient of $f$ with respect to the model parameters, and $\dho^2 f_\theta$ is the Hessian matrix of $f$.
The implicit parameter indices of the derivative tensors in \eqref{eq:gen_C} are all summed in pairs, as in the examples of \eqref{eq:ex_correlation_fns}.
Finally, the expectation value is taken over the parameter initializations, which are i.i.d. Gaussian.
% To determine how a correlation function scales as a function of $n$, one associates a \emph{cluster graph} to each such function,
\begin{defn}
    \label{def:cluster_graph} 
    Let $C(x_{1},\ldots,x_{m})$ be a correlation function with $m$ derivative tensors. We say that two derivative tensors are \emph{derivative contracted} if any of their tensor indices are summed over together in $C$. 
    The \emph{cluster graph} $G_{C}(V,E)$ of $C$ has vertices $V=\left\{ v_1,\ldots,v_m\right\}$ and edges $E=\left\{ \left(v_{i},v_{j}\right)|\text{ \ensuremath{i} is derivative contracted with \ensuremath{j}}\right\} $. 
\end{defn}
For example, the two $\dho^0 f_\theta$ derivative tensors in \eqref{eq:C20} are not derivative contracted and the two tensors in \eqref{eq:NTK} are derivative contracted.
We denote by $n_e$ ($n_o$) the number of even (odd) size components in $G_C$.
Examples of cluster graphs are shown in Figure~\ref{fig:example cluster graphs}.
\citet{dyer_asymptotics_2019} conjectured that the asymptotic behavior of a correlation function in the large width limit is $\cO(n^{n_e+(n_o-m)/2})$, allowing one to easily derive the asymptotic behavior of any correlation function from its derivative structure.

\paragraph{Our contribution.} 
The conjecture of \citet{dyer_asymptotics_2019} was proved for deep linear networks, and was shown empirically to hold for more general cases, but a complete theoretical understanding was missing.
In this work we prove the conjecture for deep networks with polynomial activations, greatly extending the regime of validity of the results derived based on the conjecture.
Our main result is the following 
\begin{theorem}
  \label{thm:main}
  Let $C(x_1,\dots,x_m)$ be a correlation function of a deep network $f_\theta : \bR^d \to \bR$ with a polynomial non-linearity $\phi$, % of rank $r=\mathcal{O}(n^0)$,
  and with $L$ hidden layers of width $n$.
  Suppose the cluster graph of $C$ has $n_e$ ($n_o$) components with an even (odd) number of vertices.
  Then
  \begin{align}
    C(x_1,\dots,x_m) = \cO(n^{n_e + (n_o - m)/2}) \,.
  \end{align}
\end{theorem}

Real analytic activation functions, such as tanh and sigmoid, can be approximated to arbitrary accuracy via a Taylor series expansion.
Our theorem applies to such activation functions if we truncate the Taylor series expansion at (any) finite order.
We verify empirically that the result holds for real-analytic and other activation functions.

\paragraph*{Related Work.}
Interest in the theoretical properties of network in the large width limit arguably originated with the work of \citet{Neal1996}, who showed that certain networks could be viewed as Gaussian processes in such limits.
See \citep{lee2017deep} for a more recent treatment in the context of deep networks, discussing their properties at initialization.
\citet{ntk} extended the analysis of such networks to the training trajectory, and showed that infinitely wide networks trained with gradient flow behave as linear models.
Follow up works extended the analysis of the training trajectory to finite width networks \citep{dyer_asymptotics_2019, huang2019dynamics}.
\citet{dyer_asymptotics_2019,littwin2020optimization} analyzed the asymptotic behavior of wide networks.
See \citet{yaida2019non, hanin2019finite, cohen2019learning} for additional theoretical results on wide networks.

\section{Theoretical results}
\label{sec:main}

In this section we prove Theorem~\ref{thm:main}, our main result.
We begin by setting up our notation and then working through several examples illustrating the methods used in the proof.

\paragraph*{Notation.}
We consider a fully-connected neural network with network map $f_\theta : \bR^d \to \bR$, $L$ hidden layers of width $n$,
and polynomial non-linearities $\phi^{(1)},\dots,\phi^{(L)}$.\footnote{
  We assume all hidden layers have the same width for simplicity.
  Our results hold in the more general case where hidden layer widths are given by $n^{(\ell)} = \alpha^{(\ell)} n$, where in the large width limit we take $n \to \infty$ while keeping the positive integers $\alpha^{(\ell)}$ fixed.
}
The post-activation of layer $\ell$ is denoted $x^{(\ell)} \in \bR^n$ and is given by
\begin{subequations}
\label{eq:genf}
\begin{align}
  x^{(1)} &:= \phi^{(1)} \left( d^{-1/2} U x \right) \,, \\
  x^{(\ell)} &:= \phi^{(\ell)} \left( n^{-1/2} W^{(\ell)} x^{(\ell-1)} \right) \,,\qquad 2<\ell\le L \,.
\end{align}
\end{subequations}
The network function is $f_\theta(x) := n^{-1/2} V^T x^{(L)}(x)$.
The model parameters are $U \in \bR^{n \times d}$, $W^{(2)},\ldots,W^{(L)} \in \bR^{n \times n}$, and $V \in \bR^n$.
We use $\theta$ to denote the collective vector of network parameters.
At initialization, each parameter is chosen i.i.d. from $\cN(0,1)$.\footnote{
  Throughout this work we will use Roman letters $i,j,k,l,\ldots$ to denote hidden layer neuron indices, going from $1$ to $n$. We will use Greek letters from the beginning of the alphabet, $\alpha,\beta,\gamma,\delta,\ldots$, to denote indices which go from $1$ to $d$, and will denote components of the input with superscripts, e.g. $x^\alpha$. We will use mid-alphabet Greek letters $\mu, \nu, \rho, \sigma$ as indices of the full parameter vector $\theta$.
  Lastly, we will use capital Roman letters $I,J,\ldots$ that run from $1$ to $m$ to index the derivative tensors in a correlation function.
}
 
Our proof relies on Isserlis' theorem, which allows one to systematically compute moments of Gaussian variables in terms of their covariance.
Suppose $X=(X_1,\ldots,X_m)$ is a centered Gaussian variable where $m$ is even.
Isserlis' Theorem states
\begin{align}
\label{eq:Isselis}
% \mathbb{E}_X\left[\prod_{I=1}^m X_I\right] = \sum_{\cP\in P_m} \prod_{(I,J)\in \cP} \mathbb{E}_X[X_I X_J],
\mathbb{E}_X\left[\prod_{I=1}^m X_I\right] &= \sum_{\cP\in P_m} \pi_{\cP}(X_1,\dots,X_m) \,,\quad
\pi_\cP(X_1,\dots,X_m) := \prod_{(I,J)\in \cP} \lexpp{X} X_I X_J \rexp \,.
\end{align}
Here $P_m$ is the collection of all partitions of $\{1,\ldots,m\}$ into pairs.
For example, if $m=4$ then
\begin{equation}
\mathbb{E}_{X}[X_{1}X_{2}X_{3}X_{4}]=\mathbb{E}_{X}[X_{1}X_{2}]\mathbb{E}_{X}[X_{3}X_4]+\mathbb{E}_{X}[X_{1}X_{3}]\mathbb{E}_{X}[X_{2}X_4]+\mathbb{E}_{X}[X_1 X_4]\mathbb{E}_{X}[X_2 X_3].\label{eq:isserlis fd}
\end{equation}

\subsection{Examples}
\label{sec:examples}

\begin{figure}
\centering
\begin{subfigure}[b]{0.15\textwidth}
    \centering
    \includegraphics[scale=0.50]{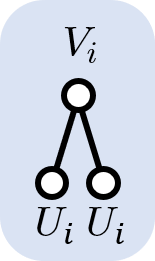}
    \caption{}
    \label{fig:example_tree_L1}
\end{subfigure}
\hfill
\begin{subfigure}[b]{0.35\textwidth}
    \centering
    \includegraphics[scale=0.50]{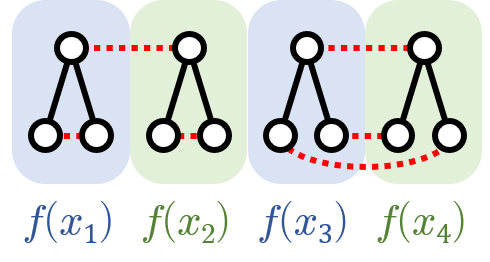}
    \caption{}
    \label{fig:example_lo_contraction}
\end{subfigure}
\hfill
\begin{subfigure}[b]{0.35\textwidth}
    \centering
    \includegraphics[scale=0.50]{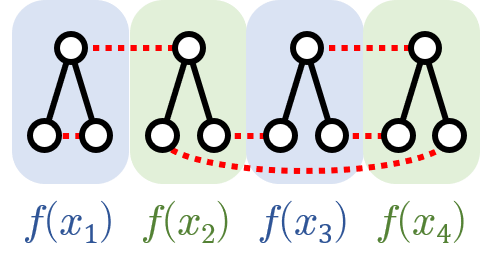}
    \caption{}
    \label{fig:example_no_contraction}
\end{subfigure}
\hfill
\caption{\textbf{(a)} A tree representing the index structure of the network map $f_\theta(x) = n^{-1/2}\sum_i V_i (U_i x)^2$, corresponding to an $L=1$ network with activation $\phi(x)=x^2$ and $d=1$. The white vertices represent indices belonging to the weights, with a black edge connecting common indices. \textbf{(b,c)} Respectively, a visual representation of two contractions contributing to $\lexpp{\theta} f_\theta(x_1) f_\theta(x_2) f_\theta(x_3) f_\theta(x_4) \rexp$. Red dotted lines connect vertices whose corresponding weights are paired in said contractions. For example, the $C_{\vec{\cP}}$ represented by the contraction (b) has the product of covariances $\lexpp{\theta}V_{i_1}V_{i_2}\rexp\lexpp{\theta}V_{i_3}V_{i_4}\rexp\lexpp{\theta}U_{i_1}U_{i_1}\rexp\lexpp{\theta}U_{i_2}U_{i_2}\rexp\lexpp{\theta}U_{i_3}U_{i_4}\rexp^2$, while for the contraction (c), it is $\lexpp{\theta}V_{i_1}V_{i_2}\rexp\lexpp{\theta}V_{i_3}V_{i_4}\rexp\lexpp{\theta}U_{i_1}U_{i_1}\rexp\lexpp{\theta}U_{i_2}U_{i_3}\rexp\lexpp{\theta}U_{i_3}U_{i_4}\rexp\lexpp{\theta}U_{i_2}U_{i_4}\rexp$. 
\label{fig:example_trees}}
\end{figure}

\paragraph{Monomial activation.}

Consider $f_\theta$ for the special case of a single hidden layer ($L=1$) having a monomial activation $\phi(x) = x^2$ and input dimension one ($d=1$) so that $f_\theta(x)=n^{-1/2}\sum_i V_i (U_i x)^2$.
It is helpful to visualize the weights $V$ and $U$ and their associated indices as having a tree-like structure, see Figure~\ref{fig:example_tree_L1}.
The correlation function consisting of rank-zero derivative tensors, $\mathbb{E}_\theta [f(x_1)f(x_2)f(x_3)f(x_4)]$, can be calculated as follows.
\begin{align}
\mathbb{E}_\theta [f(x_1)f(x_2)f(x_3)f(x_4)] &= \frac{1}{n^2}\sum_{i,j,k,l=1}^n \mathbb{E}_\theta \left[V_i V_j V_k V_l \right] 
\mathbb{E}_\theta \left[ U_i^2 U_j^2 U_k^2 U_l^2 \right]
x_1^2 x_2^2 x_3^2 x_4^2 \,. \label{eq:ex1}
\end{align}
We can now use Isserlis' theorem, with the covariance in this case determined by $\mathbb{E}_\theta [V_i V_j] = \delta_{ij}$, $\mathbb{E}_\theta [U_i U_j] = \delta_{ij}$, and $\mathbb{E}_\theta [U_i V_j] = 0$.
We will refer to a pairwise partition of the $V$ and $U$ weight factors with their indices $i,\dots,l$ unfixed as a \emph{contraction}.
A visual representation of a leading and sub-leading contraction is shown in Figures~\ref{fig:example_lo_contraction} and \ref{fig:example_no_contraction}, respectively.
Summing over these contractions, we find
\begin{align}
    \lexpp{\theta} f(x_1)f(x_2)f(x_3)f(x_4) \rexp = 3\left(9 + \frac{96}{n} \right)x_1^2 x_2^2 x_3^2 x_4^2\,.
    \label{eq:ex1-2}
\end{align}
The correlation function is $\cO(n^0)$, in agreement with Theorem~\ref{thm:main}: In this case $n_e=0$, $n_o=m=4$, and the exponent is $n_e + (n_o-m)/4=0$.
The same asymptotic behavior holds for any non-linearity $\phi(x) = x^r$ where $r$ is a non-negative integer; the case $r=1$ was studied by \citet{dyer_asymptotics_2019}.

\paragraph{Correlation function with derivatives.}
Using the same 1-hl network, consider the evaluation of the correlation function \eqref{eq:C42}. 
Using Isserlis' theorem, we find $C_{4,2}(x_1,x_2,x_3,x_4)=132 x_1^2 x_2^2 x_3^2 x_4^2 / n = \cO(n^{-1})$.
In this case we have $m=4$, $n_e = 0$, and $n_o=2$: the two odd sized clusters correspond to the two factors $\dho_{\mu_1} \dho_{\mu_2} f(x_1) \dho_{\mu_1} f(x_2) \dho_{\mu_2} f(x_3)$ and $f(x_4)$.
Theorem~\ref{thm:main} then predicts the exponent $n_e + (n_o-m)/4 = -1$, in agreement with the result above.

\paragraph{Intuition.}
These results hinge on the following relationship between derivatives of weight factors and the covariance of weight factors.
\begin{align}
  \sum_{i=1}^n \frac{\partial U_j}{\partial U_i}\frac{\partial U_k}{\partial U_i}
  & =\delta_{jk}
  = \lexpp{\theta} U_j U_k \rexp
  \, ,\qquad
  \sum_{i=1}^n \frac{\partial V_j}{\partial V_i}\frac{\partial V_k}{\partial V_i}
  =\delta_{jk}
  = \lexpp{\theta} V_j V_k \rexp
  \, . \label{eq:derivsAndIsserlis}
\end{align}
We see that a pair of derivatives with summed indices is equal to the covariance factor that was used above to compute the correlation functions.
When computing a correlation function without derivatives, Isserlis' theorem instructs us to sum over all contractions (defined by pairings of weight factors).
\citet{dyer_asymptotics_2019} showed that, for deep linear networks, computing the correlation function with a derivative pair can be done by summing over only those contractions that include the corresponding pairing.
For polynomial activations, we will show that this is still true asymptotically; derivatives acting on the post-activations can introduce additional $n$-independent coefficients.

\subsection{Main result}
\label{sec:main_proof}

In this section we describe how Isserlis' theorem can be applied to computing correlation functions with polynomial activations.
We then present the proof of Theorem~\ref{thm:main}, which is our main result.
For a polynomial activation $\phi$ of rank $r$, $f_\theta(x)$ involves two types of sums: one over polynomial terms in $\phi$ (with range $1,\dots,r$), and the other over neuron indices (with range $1,\dots,d$ for the first layer and $1,\ldots,n$ for the hidden layers).
By rearranging these two types of sums, we can express the network function in the following way.
%$f_\theta(x) = \sum_{(\vec{p},\Delta) \in \Lambda} b_{\vec{p},\Delta} f_{\vec{p},\Delta}(x)$, where
\begin{align}
  f_\theta(x) &= \sum_{(\vec{p},\Delta) \in \Lambda} b_{\vec{p},\Delta} f_{\vec{p},\Delta}(x)\, , \label{eq:ftheta_exp} \\
  f_{\vec{p},\Delta}(x) &:=
  \frac{1}{n^{(1 + p_L + \cdots + p_1)/2}}
  \sum^n_{\vec{i}^1,\vec{j}^1,\dots,\vec{i}^L}
  \Delta_{\vec{j}^{L-1},\vec{i}^{L-1},\dots,\vec{j}^1,\vec{i}^1}
  V_{i^L} \times W^{(L)}_{i^L j^{L-1}_1} \cdots W^{(L)}_{i^L j^{L-1}_{p_L}}
  \cr &\quad
  \times W^{(L-1)}_{i^{L-1}_1 j^{L-2}_1} \cdots W^{(L-1)}_{i^{L-1}_{p_{L-1}} j^{L-2}_{p_{L-1}}}
  \times \cdots \times
  W^{(2)}_{i^{2}_1 j^{1}_1} \cdots W^{(2)}_{i^{2}_{p_2} j^{1}_{p_2}} \times U_{i^1_1}(x) \cdots U_{i^1_{p_1}}(x). 
  \label{eq:fpdelta}
\end{align}
Here, $\vec{p} = (p_1,\dots,p_L)$ is a vector of non-negative integers corresponding to our choice of monomials when expanding out the terms in $\phi$, with $p_{\ell}$ the number of weight factors of type $W^{(\ell)}$ (or of type $U$ if $\ell=1$).
$\Delta$ is a product of Kronecker delta functions on pairs of indices, implementing the constraint that each element of $\vec{i}^\ell$ is equal to one of the elements of $\vec{j}^{\ell+1}$. 
$\Lambda$ is the set of all combinations $(\vec{p},\Delta)$ that are obtained by expanding out the polynomial activations.
The first layer pre-activation is written as $U_{i^1}(x) := \frac{1}{\sqrt{d}}\sum_{\alpha=1}^d U_{i^1\alpha}x_\alpha$.
Finally, $b_{\vec{p},\Delta} \in \bR$ are $n$-independent coefficients.

\begin{figure}
\centering
\begin{subfigure}[b]{0.25\textwidth}
    \centering
    \includegraphics[scale=0.50]{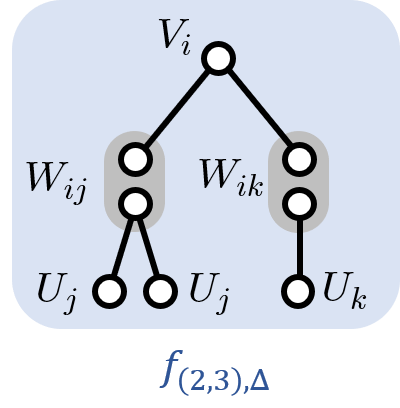}
    \caption{}
    \label{fig:example_tree}
\end{subfigure}
\hfill
\begin{subfigure}[b]{0.4\textwidth}
    \centering
    \includegraphics[scale=0.50]{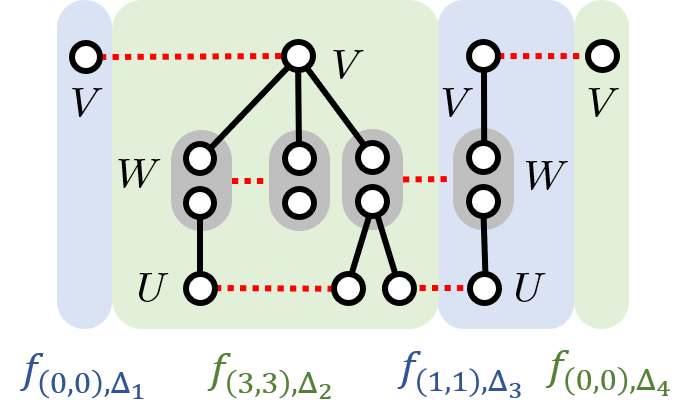}
    \caption{}
    \label{fig:example_contraction}
\end{subfigure}
\hfill
\begin{subfigure}[b]{0.25\textwidth}
    \centering
    \includegraphics[scale=0.50]{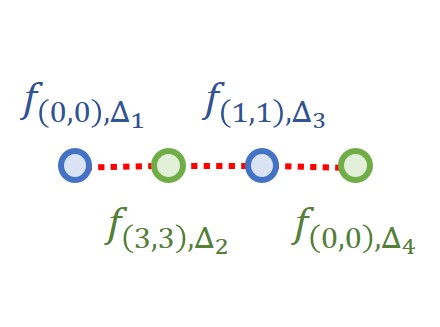}
    \caption{}
    \label{fig:example_cont_graph}
\end{subfigure}
\hfill
\caption{Graphical representation of contractions. \textbf{(a)} A forest representing the weight factors belonging to $f_{(2,3),\Delta}$ and the index structure enforced by $\Delta$ for a network of depth $L=2$. The white vertices represent nodes in the graph, each corresponding to a the index of a weight factors. Black edges represent edges, which correspond to a constraint setting the connected indices to be equal. 
The grey shaded regions group together indices that belong to the same $W$ factor.
In this example, the explicit weight factors are $V_{i^2}, W_{i^2j^1_1}, W_{i^2j^1_2}, U_{i^1_1}, U_{i^1_2}, U_{i^1_3}$ and $\Delta_{\vec{i}^1\vec{j}^1} = \delta_{j^1_1 i_1^1}\delta_{j^1_1 i_2^1} \delta_{j^1_2 i_3^1}$. \textbf{(b)} A visual representation of one of the contractions that contribute to $\lexpp{\theta} f_{(0,0),\Delta_1}f_{(3,3),\Delta_2},f_{(1,1),\Delta_3},f_{(0,0),\Delta_4}\rexp$, with red dotted lines connecting weight factors that are paired in this contraction.
\textbf{(c)} The contraction graph of the contraction shown in (b).
\label{fig:weight_trees}}
\end{figure}

The expression \eqref{eq:fpdelta} can be represented by a graph (specifically a forest), see Figure~\ref{fig:example_tree} for an example. Each node represents one index of a particular weight factor. For example, a node may represent the first index, $i_1^2$, of the weight factor $W^{(2)}_{i_1^2 j_1^1}$ in \eqref{eq:fpdelta}.
Each edge in the graph represents a constraint (a Kronecker delta factor in $\Delta$) that sets the corresponding indices to be equal.

\paragraph{Contractions.}
We are interested in bounding the asymptotic behavior of correlation functions such as $\lexpp{\theta} f(x_1) \cdots f(x_m) \rexp$.
For this purpose, it is sufficient to bound the asymptotic behavior of correlation functions of the form $ C_0(x_1,\dots,x_m) := \lexpp{\theta} f_{\vec{p}_1\Delta_1}(x_1) \cdots f_{\vec{p}_m\Delta_m}(x_m) \rexp$.
We therefore now focus on such functions, which can be written as follows.
\begin{align}
    C_0(x_1,\ldots,x_m) & =
    % \sum_{(\vec{p}_1,\Delta_1),\dots,(\vec{p}_m,\Delta_m)\in P} 
    %C_{0,}(x_1,\dots,x_m)
    % \,,\\
    % C_{0,p}(x_1,\dots,x_m) &= 
    % \lexpp{\theta} \prod_{K=1}^m
    % f_{\vec{p}_K,\Delta_K}(x_K) \rexp
    % \cr &= %%%%%%%%%%%%%%%%%%%%%%
    \frac{1}{n^{(m + \sum_{K,\ell} p_{K,\ell})/2}}
    % \sum_{\alpha}^d 
    % \left( x_{\alpha_{1,1}} \cdots x_{\alpha_{m,p_{m,1}}} \right)
    \sum_{\vec{i},\vec{j}}^n \tilde{\Delta}_{\vec{i},\vec{j}} \times \lexpp{\theta}
    V_{i^L_1} \cdots V_{i^L_m}
    \rexp 
    \cr &\quad \times
    \lexpp{\theta}
    W^{(L)}_{i^L_1 j^{(L-1)}_{1,1}} \cdots 
    W^{(L)}_{i^L_m j^{(L-1)}_{m,p_{m,L}}}
    \rexp
    \cdots
    \lexpp{\theta}
    U_{i^1_{1,1}(x_1)} \cdots
    U_{i^1_{m,p_{m,1}}(x_m)} 
    \rexp. \cr
    \label{eq:ex_by_weight}
\end{align}
Here, $\tilde{\Delta}_{\vec{i},\vec{j}} := \prod_{K=1}^m \Delta_{\vec{j}_K^{L-1},\vec{i}_K^{L-1},\dots,\vec{j}_K^1,\vec{i}_K^1}$ is a product of Kronecker delta functions, and $\vec{i},\vec{j}$ collectively represent the vectors $\vec{i}^{(\cdot)},\vec{j}^{(\cdot)}$ in \eqref{eq:fpdelta}. 
According to Isserlis' theorem, an expectation value on the right-hand side of \eqref{eq:ex_by_weight} will vanish if it includes an odd number of weight factors.
We can therefore ignore such terms as they do not affect the asymptotic behavior, and from now on we will only consider terms with an even number of weight factors in each layer.
Using Isserlis' theorem, we can express these expectation values as follows.
\begin{align}
    C_0(x_1,\dots,x_m) &= 
    \sum_{\vec{\cP}\in P} C_{\vec{\cP}}(x_1,\dots,x_m)
    \,, \label{eq:C0_contractions} \\
    C_{\vec{\cP}}(x_1,\dots,x_m) &:=
    \frac{1}{n^{(m + \sum_{k,\ell} p_{k,\ell})/2}}
    % \sum_{\alpha}^d 
    % \left( x_{\alpha_{1,1}} \cdots x_{\alpha_{m,p_{m,1}}} \right)
    \sum_{\vec{i},\vec{j}}^n \tilde{\Delta}_{\vec{i},\vec{j}}
    \pi_{\cP_{L+1}} \left( V_{i^L_1}, \dots, V_{i^L_m} \right)
    \cr &\quad \times
    \pi_{\cP_L}\left(
    W^{(L)}_{i^L_1 j^{(L-1)}_{1,1}}, \dots,
    W^{(L)}_{i^L_m j^{(L-1)}_{m,p_{m,L}}}
    \right)
    \cdots
    \pi_{\cP_1}\left(
    U_{i^1_{1,1}}(x_1), \dots,
    U_{i^1_{m,p_{m,1}}}(x_m) \right) \,. \cr
    \label{eq:Cp}
\end{align}
Here $\vec{\cP}$ is a vector, where each element is a partition of the relevant number of weight factors into pairs, as explained in Isserlis' theorem (c.f. \eqref{eq:Isselis}).
The set of all such pairings is denoted by $P$.
We will call each vector of pairings $\vec{\cP}$ a \emph{contraction}.
Figure~\ref{fig:example_contraction} shows a contraction graphically: Each dashed line corresponds to a pairing of the corresponding weight factors in $\vec{\cP}$.\footnote{
$\cP_\ell$ is a partition of the $\sum_{k=1}^m p_{k,\ell}$ weight factors that are the arguments of $\Gamma_{\cP_\ell}(\cdot)$ and $p_{k,L+1}:=1$.
}
The contribution of each contraction to the correlation function is given by $C_{\vec{\cP}}$.

To each contraction $\vec{\cP}$ we associate a \emph{contraction graph} $\Gamma_{\vec{\cP}}(v,e)$,
with vertices $v=\{1,\dots,m\}$ corresponding to the factors $f_K := f_{\vec{p_K},\Delta_K}(x_K)$, $K=1,\dots,m$ in the correlation function.
There is an edge between two nodes $K_1, K_2 \in v$ if the contraction $\vec{\cP}$ includes a pairing between weight factors belonging to $f_{K_1},f_{K_2}$.
Figure~\ref{fig:example_cont_graph} shows an example of a contraction graph.
The following result establishes a relationship between contraction graphs and the cluster graph of the corresponding correlation function.

\begin{lemma}
Any correlation function can be written as  $C(x_1,\dots,x_m) = \sum_{\vec{\cP} \in P} a_{\vec{\cP}} C_{\vec{\cP}}(x_1,\dots,x_m)$ where $C_{\vec{\cP}}$ are functions of the form \eqref{eq:Cp}, $a_{\vec{\cP}}$ are $n$-independent non-zero coefficients, and the sum is over a set $P$ of contractions. 
% \KA{Do we need to specify the exact relation between $C$ and the $C_{\vec{\cP}}$? I.e. the $C_{\vec{\cP}}$ are those that contribute to the same correlation function $C$, with all explicit derivatives removed.}
Let $\vec{\cP} \in P$ be a contraction, whose contraction graph has $N$ components.
Let $n_e$ ($n_o$) be the number of even (odd) components in the cluster graph of the correlation function $C$.
Then $N \le n_e + \frac{n_o}{2}$.
\label{lem:main}
\end{lemma}
Let us describe the intuition behind this result.
As shown in \eqref{eq:derivsAndIsserlis}, a pair of derivatives in a correlation function always leads to a Kronecker delta factor between indices of the corresponding function factors.
In other words, every pair of derivatives that appears in a correlation function leads to a particular pairing between factors that shows up in every contraction.
This restricts the set of contractions that contribute to the correlation function, as expressed in the lemma.
The proof of Lemma~\ref{lem:main} can be found in \swaptext{the SM}{Appendix \ref{app:gen_deriv_tensor}}.
We now turn to the proof of the main theorem.

\begin{proof}[Proof (Theorem~\ref{thm:main}).] 
From Lemma~\ref{lem:main}, the correlation function can be written as a sum over contractions, $C(x_1,\dots,x_m) = \sum_{\vec{\cP} \in P} a_{\vec{\cP}} C_{\vec{\cP}}(x_1,\dots,x_m)$.
Let $\vec{\cP} \in P$ be a contraction, and let $\Gamma_{\vec{{\cP}}}$ be its contraction graph with $N$ components.
Below we will show that $C_{\vec{\cP}}(x_1,\dots,x_m) = \cO(n^{N-m/2})$.
It then follows from Lemma~\ref{lem:main} that $C(x_1,\dots,x_m) = \cO(n^{N-m/2}) = \cO(n^{n_e + (n_o-m)/2})$, concluding the proof.

It is left to show that $C_{\vec{P}} = \cO(n^{N - m/2})$.
We will assume that the contraction has an even number of weight factors at each layer, because otherwise the contribution to the correlation function vanishes due to Isserlis' theorem.
We will now proceed by induction over the number of pairings in the contraction.
For the induction base, we take a contraction with $0$ pairings has $N=0$, $m=0$, and value $1$.
For the induction step, we first assume that $C_{\vec{\cP}}(x_1,\dots,x_m) = \cO(n^{N-m/2})$.
We then add two weight factors of the same layer to the contraction (keeping the number of weights at each layer even), and pair them.

Notice that any contraction of the form \eqref{eq:C0_contractions} can be obtained by adding pairs in this way one at a time.
To see this, one can start with a given contraction and keep removing pairs one at a time from the lower-most available layers until all the weight factors are gone.
Then reverse the order to build the contraction from scratch.
We denote the revised contraction by $\vec{\cP}'$, with $m'$ vertices and $N'$ components in its contraction graph.
The contribution $C_{\vec{\cP}'}$ includes these two additional weight factors, and possible $n$ sums and $n^{-1/2}$ factors as needed to maintain the form \eqref{eq:fpdelta}.
In particular, the indices of the added pair are constrained by a revised $\tilde{\Delta}$ as needed.\footnote{
For example, the constraint implies that we can only add a pair of $U$ weights if there is at least one $W^{(2)}$ factor (assuming $L>1$), because the $U$ index must be set to equal to the $j$ index of some $W^{(2)}_{ij}$ factor.
}

In order to bound $C_{\vec{\cP}'}$ we need to keep track of the factors that affect the large width asymptotics.
When adding a weight factor, it adds one or two $n$ sums to $C_{\vec{\cP}'}$ (by this we mean a sum over an index that goes from $1,\dots,n$, as well as an explicit $n^{-1/2}$ normalization factor.
A pairing of two weight factors in a contraction leads to one or two Kronecker deltas, each of which can potentially turn a double $n$ sum into a single sum.
The asymptotic behavior then follows from the final number of $n$ sums (after accounting for the Kronecker deltas), and from the explicit $n^{-1/2}$ factors.
We now consider how these are affected for every type of weight factor pair we can add.

\textbf{Add a $V$ pair.} 
Every function factor $f_{\vec{p},\Delta}(x)$ in the correlation function has a single $V$ factor, so adding a $V$ pair changes $m' = m+2$.
We also have $N' = N + 1$ because the added vertices belong to the same component.
Therefore, $N' - \frac{m'}{2} = N - \frac{m}{2}$.
The asymptotic behavior does not change, because $C_{\vec{\cP}'} = C_{\vec{\cP}} \cdot \frac{1}{n} \sum_{i,j}^n \lexpp{\theta} V_i V_j \rexp = n^{-1} \sum_{i,j} \delta_{ij} C_{\vec{\cP}} = C_{\vec{\cP}}$.
We added two indices (two $n$ sums), a single Kronecker delta, and two $n^{-1/2}$ factors, and these contributions cancel.

\begin{table}
\centering
\begin{tabular}{c|cc|ccc|ccc}
  $C$ & $(n_e,n_o)$ & pred. exp. & tanh & sigmoid & softplus & linear & relu & hard-sigmoid \\ %& ELU\\
\hline
$C_{2,0}$ & $(0,2)$ & $0$  & $-0.01$ & $0.01$ & $-0.03$ & $0.00$ & $-0.01$ & $-0.03$ \\ %& $-0.04$ \\
$C_{2,1}$ & $(1,0)$ & $0$  & $0.00$ & $0.00$ & $-0.00$ & $0.00$ & $-0.00$ & $-0.00$ \\ %& $-0.02$ \\
$C_{4,0}$ & $(0,4)$ & $0$  & $-0.04$ & $0.03$ & $-0.03$ & $0.01$ & $-0.05$ & $-0.00$ \\ %& $0.30$ \\
$C_{4,2}$ & $(0,2)$ & $-1$  & $-1.03$ & $-1.00$ & $-1.02$ & $-1.02$ & $-1.01$ & $-1.00$ \\ %& $-0.98$ \\
$C_{4,3}$ & $(1,0)$ & $-1$  & $-0.99$ & $-1.01$ & $-1.01$ & $\mathbf{-2.03}$ & $\mathbf{-2.00}$ & $\mathbf{-2.01}$ \\ %& $-1.00$ \\
$C_{6,4}$ & $(0,2)$ & $-2$  & $-1.96$ & $-2.02$ & $-1.99$ & $-2.00$ & $-1.99$ & $-2.00$ \\ %& $-2.03$ 
\end{tabular}
\medskip
\caption{Numerical results measuring the asymptotic behavior of different correlation functions at large width.
Here, $C$ are the correlation functions as defined in \eqref{eq:ex_correlation_fns} and \eqref{eq:numcorrs}, $(n_e,n_o)$ are the number of even and odd components in the correlation function's cluster graph, and pred. exp. is the exponent, $n_e+(n_o-m)/2$, predicted by Theorem~\ref{thm:main}.
The six right-most columns list the measured exponents for different activation functions.
Each experiment (with fixed correlation function and activation function) involved a fully-connected networks with 3 hidden layers and width ranging between $2^7,\dots,2^{13}$.
For each width, the correlation function was estimated by averaging over $10^3$ initializations.
The exponent was then measured by fitting a power law to the width dependence.
\label{tab:smooth}}
\end{table}

\textbf{Add a $U$ pair.} 
Here we have $m'=m$ and either $N'=N$ or $N'=N-1$, where the latter holds if the new pairing connects two separate components.
First, suppose that $N'=N$.
Each $U(x)$ factor introduces an $n$ sum.
Each factor also introduces a Kronecker delta, setting its $n$ index to an existing weight index from a previous layer (this is part of the $\tilde{\Delta}$ constraint).
Overall, no $n$ sums are added in this case, and the asymptotic behavior does not become more divergent.
Therefore, $C_{\cP'} = \cO(n^{N-m/2}) = \cO(n^{N'-m'/2})$.
Now, suppose that $N' = N-1$.
In this case, the new pair connects two components in the contraction graph that were previously disconnected.
Therefore, $C_{\vec{\cP}}$ does not have any Kronecker deltas equating the indices of weight factors belonging to these two components.
As a result, the Kronecker delta from the new $U$ pairing equates two indices that were previously unrelated, reducing the number of $n$ sums by 1 (going from $C_{\vec{\cP}}$ to $C_{\vec{\cP}'}$).
We then have $C_{\vec{\cP}'} = \cO(n^{-1} C_{\vec{\cP}}) = \cO(n^{N-1-m/2}) = \cO(n^{N'-m'/2})$ (the $n^{-1}$ is due to the explicit normalization factors).

\textbf{Add a $W^{(\ell)}$ pair.} 
Again, when adding this pair we have $m'=m$, and either $N'=N$ or $N'=N-1$, because adding a single edge to the contraction graph cannot reduce the number of components further. 
Suppose the pair leaves $N$ unchanged.
The new pair introduces a $n^{-1}$ factor in $C_{\vec{\cP}'}$, as well as two new $n$ sums, over the second index of each $W^{(\ell)}$ factor.
The first index of each factor does not introduce new sums because of the $\tilde{\Delta}$ constraint, which sets it equal to another index from a previous layer.
The pairing introduces at least one new delta function connecting these second indices, which combines the two sums into one.
Therefore, after introducing this pairing we have an additional $n$ sum and an additional $n^{-1}$ explicit normalization factor, resulting in $C_{\vec{\cP}'} = \cO(n^{-1} n \vec{\cP}) = \cO(n^{N-m/2})$.
Now, suppose that $N' = N-1$.
As in the $U$ case, this implies that the new pairing connects previously-separated components, and therefore the corresponding Kronecker delta added to $C_{\vec{\cP}'}$ will reduce the number of $n$ sums by 1.
We find that $C_{\vec{\cP}'} = \cO(n^{-1} C_{\vec{\cP}}) = \cO(n^{N-1-m/2}) = \cO(n^{N'-m'/2})$.
\end{proof}

\section{Numerical experiments}
\label{sec:numerics}

We now present numerical results measuring the asymptotic behavior of the correlation functions defined in \eqref{eq:ex_correlation_fns}, and also the following correlation functions.
\begin{subequations}
\begin{align}
  C_{4,0}(x_1,...,x_4) &:= \lexpp{\theta} f(x_1) f(x_2) f(x_3) f(x_4) \rexp \,, \\
  C_{4,3}(x_1,...,x_4) &:= \sum_{\mu_1,\mu_2,\mu_3} \lexpp{\theta}
  \dho_{\mu_1} \dho_{\mu_3} \dho_{\mu_3} f(x_1) \dho_{\mu_1} f(x_2) \dho_{\mu_2} f(x_3) \dho_{\mu_3} f(x_4) \rexp
  \,, \label{eq:C43} \\
  C_{6,4}(x_1,...,x_6) &:= \sum_{\mu_1,\dots,\mu_4}
  \lexpp{\theta} 
  \dho_{\mu_1} \dho_{\mu_2} f(x_1) 
  \dho_{\mu_1} f(x_2) 
  \dho_{\mu_2} f(x_3) 
  \dho_{\mu_3} \dho_{\mu_4} f(x_4) 
  \dho_{\mu_3} f(x_5) 
  \dho_{\mu_4} f(x_6) 
  \rexp.
\end{align}
\label{eq:numcorrs}
\end{subequations}
Table~\ref{tab:smooth} lists the measured exponents for various activation functions, compared against the theoretical prediction.
In all cases, we find that the conjecture of \citet{dyer_asymptotics_2019} holds.
We note a difference between the asymptotic behavior for two classes of activation functions.
For networks with non-linear, real-analytic activations, the upper bound predicted by the conjecture is tight.
For networks with piecewise-linear activations the bound always holds but is not always tight; see in particular the correlation function $C_{4,3}$.\footnote{
As shown by \citet{dyer_asymptotics_2019}, for a deep linear network the exponent can be derived using a Feynman diagram calculation.
Indeed, a full Feynman diagram calculation of $C_{4,3}$ leads to the answer -2 for the exponent, in agreement with the measured value.
% The leading diagram in this case has 2 boundaries instead of 1, which explains the additional suppression.
  }
In our experiments, the asymptotic behavior of networks with piecewise-linear activations matches that of deep linear networks.
Explicit calculations for networks with polynomial activations show correlation functions such as $C_{4,3}$ have additional contributions that would vanish in the linear case.
Said additional contributions can be a higher-order and hence lead to different asymptotic beavhior. 
See \swaptext{the SM}{Appendix~\ref{app:pwl}} for further discussion.

\section{Discussion}
\label{sec:conclusion}

We build on the work of \citet{dyer_asymptotics_2019}, who presented a conjecture regarding the asymptotic behavior of wide neural networks.
The conjecture is useful in the study of network dynamics.
Among other results, it allows one to go beyond infinite width results, and analytically derive the gradient descent trajectory of networks with large but finite width.
It is therefore of interest to prove the general form of the conjecture, which was previously established for the special case of deep linear networks.
In this paper we prove the conjecture for networks with polynomial activations, greatly extending its validity.%\footnote{We also use Theorem \ref{conj:ethan guy conj fd} to bound the covariance of two sets of contracted derivative tensors, see \swaptext{the SM}{Appendix \ref{app:Cov_asym}} for details.}

Real-analytic activation functions such as $\tanh$ and sigmoid can be approximated to arbitrary precision via a Taylor series expansion.
Therefore, our theorem is applicable if one truncates such expansions in said networks at some finite order.
We verify the validity of the conjecture empirically for a variety of correlation functions and activation functions.
We point out a difference between the asymptotic behavior of networks with non-linear, real-analytic activations functions, vs. the behavior of networks with piecewise-linear activations such as ReLU.
In the real-analytic case, the asymptotic bound predicted by the conjecture is tight, while in the piecewise-linear case the bound is sometimes not tight.
Instead, the asymptotics of such networks agree with those of the deep linear networks analyzed in \citet{dyer_asymptotics_2019}.
See the \swaptext{SM}{Appendix} for an initial investigation of the difference between these two classes of activation functions.

\swaptext{
\section*{Broader impact}
This work does not present any foreseeable societal consequence.
}
{
\section*{Acknowledgments}
The work of KA was supported, in part, by the Simons Foundation as part of the Simons Collaboration on Ultra Quantum Matter. KA would like to thank Andreas Karch for useful discussions. GG would like to thank Ethan Dyer, Michael Douglas, Yasaman Bahri, Boris Hanin, and Jascha Sohl-Disckstein for useful discussions.
}

\bibliographystyle{plainnat}
\bibliography{wide_networks}

\begin{thebibliography}{13}
\providecommand{\natexlab}[1]{#1}
\providecommand{\url}[1]{\texttt{#1}}
\expandafter\ifx\csname urlstyle\endcsname\relax
  \providecommand{\doi}[1]{doi: #1}\else
  \providecommand{\doi}{doi: \begingroup \urlstyle{rm}\Url}\fi

\bibitem[Belkin et~al.(2018)Belkin, Hsu, Ma, and Mandal]{belkin2018reconciling}
Mikhail Belkin, Daniel Hsu, Siyuan Ma, and Soumik Mandal.
\newblock Reconciling modern machine learning practice and the bias-variance
  trade-off.
\newblock \emph{arXiv preprint arXiv:1812.11118}, 2018.

\bibitem[Cohen et~al.(2019)Cohen, Malka, and Ringel]{cohen2019learning}
Omry Cohen, Or~Malka, and Zohar Ringel.
\newblock Learning curves for deep neural networks: A gaussian field theory
  perspective.
\newblock \emph{arXiv preprint arXiv:1906.05301}, 2019.

\bibitem[Dyer and Gur-Ari(2019)]{dyer_asymptotics_2019}
Ethan Dyer and Guy Gur-Ari.
\newblock Asymptotics of wide networks from feynman diagrams.
\newblock \emph{arXiv preprint arXiv:1909.11304}, 2019.

\bibitem[Hanin and Nica(2019)]{hanin2019finite}
Boris Hanin and Mihai Nica.
\newblock Finite depth and width corrections to the neural tangent kernel.
\newblock \emph{arXiv preprint arXiv:1909.05989}, 2019.

\bibitem[Huang and Yau(2019)]{huang2019dynamics}
Jiaoyang Huang and Horng-Tzer Yau.
\newblock Dynamics of deep neural networks and neural tangent hierarchy, 2019.

\bibitem[{Jacot} et~al.(2018){Jacot}, {Gabriel}, and {Hongler}]{ntk}
Arthur {Jacot}, Franck {Gabriel}, and Cl{\'e}ment {Hongler}.
\newblock {Neural Tangent Kernel: Convergence and Generalization in Neural
  Networks}.
\newblock \emph{arXiv e-prints}, art. arXiv:1806.07572, June 2018.

\bibitem[Lee et~al.(2017)Lee, Bahri, Novak, Schoenholz, Pennington, and
  Sohl-Dickstein]{lee2017deep}
Jaehoon Lee, Yasaman Bahri, Roman Novak, Samuel~S Schoenholz, Jeffrey
  Pennington, and Jascha Sohl-Dickstein.
\newblock Deep neural networks as gaussian processes.
\newblock \emph{arXiv preprint arXiv:1711.00165}, 2017.

\bibitem[{Lee} et~al.(2019){Lee}, {Xiao}, {Schoenholz}, {Bahri},
  {Sohl-Dickstein}, and {Pennington}]{2019arXiv190206720L}
Jaehoon {Lee}, Lechao {Xiao}, Samuel~S. {Schoenholz}, Yasaman {Bahri}, Jascha
  {Sohl-Dickstein}, and Jeffrey {Pennington}.
\newblock {Wide Neural Networks of Any Depth Evolve as Linear Models Under
  Gradient Descent}.
\newblock \emph{arXiv e-prints}, art. arXiv:1902.06720, Feb 2019.

\bibitem[Littwin et~al.(2020)Littwin, Galanti, and
  Wolf]{littwin2020optimization}
Etai Littwin, Tomer Galanti, and Lior Wolf.
\newblock On the optimization dynamics of wide hypernetworks.
\newblock \emph{arXiv preprint arXiv:2003.12193}, 2020.

\bibitem[Neal(1996)]{Neal1996}
Radford~M. Neal.
\newblock \emph{Priors for Infinite Networks}, pages 29--53.
\newblock Springer New York, New York, NY, 1996.
\newblock ISBN 978-1-4612-0745-0.
\newblock \doi{10.1007/978-1-4612-0745-0\_2}.
\newblock URL \url{https://doi.org/10.1007/978-1-4612-0745-0\_2}.

\bibitem[Neyshabur et~al.(2017)Neyshabur, Bhojanapalli, McAllester, and
  Srebro]{neyshabur2017exploring}
Behnam Neyshabur, Srinadh Bhojanapalli, David McAllester, and Nati Srebro.
\newblock Exploring generalization in deep learning.
\newblock In \emph{Advances in Neural Information Processing Systems}, pages
  5947--5956, 2017.

\bibitem[Yaida(2019)]{yaida2019non}
Sho Yaida.
\newblock Non-gaussian processes and neural networks at finite widths.
\newblock \emph{arXiv preprint arXiv:1910.00019}, 2019.

\bibitem[Zagoruyko and Komodakis(2016)]{zagoruyko2016wide}
Sergey Zagoruyko and Nikos Komodakis.
\newblock Wide residual networks.
\newblock \emph{arXiv preprint arXiv:1605.07146}, 2016.

\end{thebibliography}

\newpage

\begin{appendix}

\swaptext{\section*{Supplementary material}}{\section*{Appendix}}

\section{Review of previous work}
\label{app:review}
In this appendix we briefly review some results of \citet{dyer_asymptotics_2019}. 
This includes the statement of their main conjecture as well as a brief overview of how calculations were done for linear networks. 

As we have mentioned in the main text, the primary result of \citet{dyer_asymptotics_2019} is a conjecture that relates a correlation function's asymptotic behavior to the properties of its cluster graph (see Definition~\ref{def:cluster_graph}). 
The conjecture states,
\begin{conj} 
\label{conj:ethan guy conj fd} Consider a correlation function $C(x_1,\ldots,x_m)$ with cluster graph $G_C$ that has $n_e$ and $n_o$ even and odd components, respectively. The asymptotic behavior of $C$ is
\begin{equation}
C(x_1,\ldots,x_m)=\mathcal{O}(n^{n_e+(n_o-m)/2}) \, .
\end{equation}
\end{conj}
This conjecture was proven for the case in which the network $f_\theta$ was linear, i.e. the activation functions $\phi$ in \eqref{eq:genf} are all the identity. 
A proof was also shown for the special cases of $f_\theta$ being a single hidden-layer ($L=1$) non-linear network, which we review in Appendix \ref{sec:single_nonlin} below, as well as the case where all $\phi$ are ReLU activation functions and all inputs to the network are equal \citep{dyer_asymptotics_2019}. 
Although these cases exclude the most general form of \eqref{eq:genf}, numerical results seem to support this bound for various deeper non-linear networks. 
For the most part the bound was tight, but occasionally one would observe a difference in asymptotic behavior between piece-wise linear activations (e.g. linear or ReLU) and real analytic activations (e.g. $\tanh$ or softplus).

\subsection{Linear networks}

Here we briefly demonstrate how \citet{dyer_asymptotics_2019} calculated correlation functions for linear networks. 
This also serves as a pedagogical introduction to methods used to evaluate correlation functions throughout the main text.

A one hidden-layer linear network with $d=1$ is given by $f_\theta(x)=\frac{1}{\sqrt{n}}\sum_{i} V_i U_{i}x$. 
Consider the simplest non-trivial correlation function, $\lexpp{\theta}f(x_1)f(x_2)\rexp$. 
In the linear case we can evaluate the correlation function using Isserlis' Theorem, see \eqref{eq:Isselis} in the main text.
Isserlis' Theorem can be applied directly to the Gaussian distributed weights of the network after writing them out explicitly,
\begin{align}
\label{eq:linear_one}
\lexpp{\theta}f(x_1)f(x_2)\rexp & =\frac{1}{n}\sum_{i,j=1}^n \lexpp{\theta}V_i V_j\rexp\lexpp{\theta}U_i U_j \rexp x_1 x_2= x_1 x_2= \mathcal{O}(n^0)\,.
\end{align}

This technique is straightforward to generalize to other linear network correlation functions with rank-zero derivative tensors and supports Conjecture~\ref{conj:ethan guy conj fd}. 
For example, the $m=4$ case is given by
\begin{align}
\lexpp{\theta}f(x_1)f(x_2)f(x_3)f(x_4)\rexp & = \frac{1}{n^2}\sum_{i,j,k,l}^n \lexpp{\theta} V_i V_j V_k V_l \rexp \lexpp{\theta} U_i U_j U_k U_l \rexp x_1 x_2 x_3 x_4 \, , \nonumber \\
& =\frac{1}{n^2}\sum_{i,j,k,l}^n  \left(\delta_{ij}\delta_{kl}+\delta_{ik}\delta_{jl}+\delta_{il}\delta_{jk}\right) \nonumber \\
& \quad \qquad \times\left(\delta_{ij}\delta_{kl}+\delta_{ik}\delta_{jl}+\delta_{il}\delta_{jk}\right)x_1 x_2 x_3 x_4 \, ,\nonumber \\
& = 3\left(1+\frac{2}{n}\right)x_1 x_2 x_3 x_4 \,,
\end{align}
and is thus $\mathcal{O}(n^0)$. 
It is worth noting that, of the nine possible terms in the second equality, it is the three terms which have contraction structure $\lexpp{\theta}V_i V_j\rexp \lexpp{\theta}U_i U_j\rexp$ that produce the leading-order terms (as opposed to, say, terms like $\lexpp{\theta}V_i V_j\rexp \lexpp{\theta}U_k U_k\rexp$). 
That is, when the weights belonging to two derivative tensors are contracted pairwise, we produce the leading-order asymptotic behavior.
This will be a general trend in what follows and continues to be true in the non-linear polynomial activations we examine in the main text.

\paragraph*{Adding Derivatives}

It is also straightforward to evaluate correlation functions with derivatives in the linear case. 
First we note that taking a derivative with respect to a network weight effectively reduces the number of weights present in the derivative tensor. 
For example, in the same network considered above,
\begin{equation}
\frac{\partial f(x)}{\partial U_i}=\frac{1}{\sqrt{n}}\sum_{j=1}^n \frac{\partial}{\partial U_i}\left(V_j U_j x \right)= \frac{1}{\sqrt{n}}\sum_{j=1}^n \delta_{ij} V_j x \,.
\end{equation}
An immediate consequence of this is that taking multiple derivatives with respect to the \emph{same} type of weight vanishes,
\begin{equation}
\frac{\partial^2 f(x)}{\partial U_i \partial U_j }=\frac{1}{\sqrt{n}}\frac{\partial}{\partial U_{j}}\left(V_i x\right) = 0 \,.
\end{equation}
% This will be an important difference from the nonlinear case that we highlight in Section~\ref{sec:pwl_vs_smooth} of the main text.

As an explicit example, for a single hidden layer with $d=1$, our set of weights consist of $\theta^{\mu}\in\left\{ V_{i},U_{i}\right\}_{i=1,\ldots,n}$ and so the NTK \eqref{eq:NTK} is
\begin{equation}
\sum_\mu \mathbb{E}_{\theta}\left[\frac{\partial f(x_1)}{\partial\theta^{\mu}}\frac{\partial f(x_2)}{\partial\theta^{\mu}}\right]=\sum_{i=1}^n \lexpp{\theta}\frac{\partial f(x_1)}{\partial U_i}\frac{\partial f(x_2)}{\partial U_i}+\frac{\partial f(x_1)}{\partial V_i}\frac{\partial f(x_2)}{\partial V_i}\rexp \, .
\end{equation}
Each of these terms can still be evaluated using Isserlis' Theorem, each derivatives simply removes one weight,
\begin{subequations}
\label{eq:ntk_linear}
\begin{align}
\sum_{i=1}^n \lexpp{\theta}\frac{\partial f(x_1)}{\partial V_{i}}\frac{\partial f(x_2)}{\partial V_i} \rexp & =\frac{1}{n} \sum_{i,j,k}^n \lexpp{\theta}\delta_{ij}\delta_{ik}U_j U_k \rexp x_1 x_2 = x_1 x_2 = \mathcal{O}(n^0)\, ,\label{eq:linear deriv 1 fd}\\
\sum_{i=1}^n \lexpp{\theta}\frac{\partial f(x_1)}{\partial U_i}\frac{\partial f(x_2)}{\partial U_i}\rexp & =\frac{1}{n} \sum_{i,j,k}^n\lexpp{\theta}V_j V_k \delta_{ij}\delta_{ik}\rexp x_1 x_2 =  x_1  x_2 = \mathcal{O}(n^0)\, .
\end{align}
\end{subequations}
Thus we conclude the NTK scales as $\mathcal{O}(n^0)$ in the linear case. 

As we discussed in Section~\ref{sec:main}, we see that the calculation and results of \eqref{eq:linear_one} and \eqref{eq:ntk_linear} are quite similar. 
This lead \citet{dyer_asymptotics_2019} to propose a method of keeping track of derivative contractions by forcing a certain subset of contractions. 
This is extended to the polynomial case in Lemma~\ref{lem:dt} of Appendix~\ref{app:gen_deriv_tensor}. 
Using this method, one can generalize the method above to arbitrary correlation functions, which leads to the bound of Conjecture~\ref{conj:ethan guy conj fd} \citep{dyer_asymptotics_2019}.

\subsection{Sketch of 2-layer proof}
\label{sec:single_nonlin}

For the special case of a single hidden-layer network, \citet{dyer_asymptotics_2019} proposed a way to calculate the asymptotic behavior of correlation functions with non-linearities.
We sketch the methodology of this proof here, mainly to highlight its distinction from our methods in the main text. 
As an example, once again consider the $L=1$ and $d=1$ network with network functional $f_\theta(x)=\frac{1}{\sqrt{n}}\sum_{i} V_i \phi\big(U_i x \big)$, in which case the correlation function in \eqref{eq:linear_one} is now
\begin{align}
\lexpp{\theta}f(x_1)f(x_2)\rexp & =\frac{1}{n}\sum_{i,j=1}^n \lexpp{\theta}V_{i}V_{j}\rexp\lexpp{\theta}\phi\left( U_i x_1 \right)\phi\left(U_j x_2 \right)\rexp \, , \nonumber \\ 
& =\frac{1}{n}\sum_{i=1}^n \lexpp{\theta}\phi\left( U_i x_1 \right)\phi\left(U_i x_2 \right)\rexp \,.
\end{align}
In the last term, everything in the $\mathbb{E}_{\theta}[\cdots]$ is independent of $n$, and so although we cannot evaluate the exact expectation, it must be $\cO(n^0)$.
Since we are summing over $n$ such terms, we can conclude
\begin{equation}
\label{eq:one hidden ff}
\mathbb{E}_{\theta}\left[f(x_1)f(x_2)\right]=\mathcal{O}(n^0)\,.
\end{equation}
Thus we find the same asymptotic behavior as the linear case. 
This technique can be generalized to other $L=1$ correlation functions because one can take advantage of the fact that the $V$-weights are always outside the activations. 
Unfortunately, this techniques does not generalize well to finding the asymptotic behavior for arbitrary depth nonlinear networks because we cannot use Isserlis' Theorem on the non-linearities, whose outputs are not in general Gaussian.

\section{Theoretical results}
\label{app:gen_deriv_tensor}

In this appendix we present the proofs of key results, including Lemma~\ref{lem:main}.
We begin by introducing additional notation, generalizing the discussion of the correlation function $C_0$ in the theory section, \eqref{eq:ex_by_weight}, to the case of correlation functions with derivatives.

The most general correlation function with $m$ arbitrary rank derivative tensors is that of \eqref{eq:gen_C}, reproduced here for convenience, $\sum_{\rm indices} \lexpp{\theta} \partial^{k_1}f_\theta(x_1)\cdots\partial^{k_m}f_\theta(x_m)\rexp$.
We again assume all such derivatives are summed in pairs over the model parameters, which is what is meant by the ``$\sum_{\text{indices}}$''.
To begin, we would like to generalize \eqref{eq:fpdelta} to the case of a rank-$k$ derivative tensor. 
Define $F_{\vec{p},\Delta}^k := \partial^{k} f_{\vec{p}, \Delta}$, which has $k$ derivatives with respect to weights acting on $f_{\vec{p},\Delta}$, with the $k=0$ case simply $F_{\vec{p},\Delta}^0 := f_{\vec{p}, \Delta}$. 
Then, generalizing \eqref{eq:ftheta_exp}, we can write
\begin{align}
    \partial^k f_\theta(x) &= \sum_{(\vec{p},\Delta) \in \Lambda} b_{\vec{p},\Delta} F^k_{\vec{p},\Delta}(x)\,,
\end{align}
where $b_{\vec{p},\Delta}$ is an $n$-independent constant.
As with the correlation containing only rank-zero derivative tensors, to bound the asymptotic behavior of the most-general correlation function, \eqref{eq:gen_C}, it is enough to bound the asymptotic behavior of correlations of the form
\begin{align}
  \tilde{C}(x_1,\ldots,x_m) = \sum_{\rm indices} \lexpp{\theta} \prod_{K=1}^m F_K \rexp\, .
  \label{eq:GenCterm}
\end{align}
Here and throughout this appendix we will keep the parameters of a given $F$ implicit, i.e. $F_K:=F_{\vec{p}_K, \Delta_K}^{k_K}(x_K)$.

The presence of derivatives with respect to weights in this correlation function does not change the fact that one can evaluate it with Isserlis' theorem.
The derivatives simply serve to remove certain weights from expectations, but introduce contraction-dependent coefficients $a_{\vec{\cP}}$ (see examples in Section~\ref{sec:examples}7). 
We will show we can still write \eqref{eq:GenCterm} as a sum over contractions, and so the analog of \eqref{eq:C0_contractions} is 
\begin{align}
    \tilde{C}(x_1,\ldots,x_m) = \sum_{\vec{\cP}\in P} a_{\vec{\cP}}C_{\vec{\cP}}(x_1,\ldots,x_m) \,,
    \label{eq:derive_contractions}
\end{align}
where the coefficients $a_{\vec{\cP}}$ are $n$-independent, and hence to find the asymptotic scaling of \eqref{eq:GenCterm} it is enough to find the asymptotic scaling of the $C_{\vec{\cP}}$.

\begin{figure}
\centering
\begin{subfigure}[b]{0.45\textwidth}
    \centering
    \includegraphics[scale=0.65]{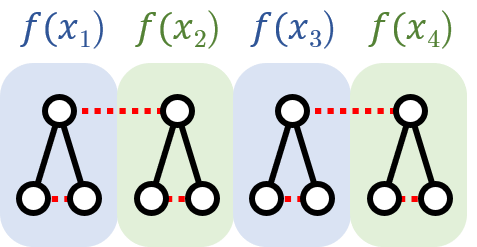}
    \caption{}
\end{subfigure}
\hfill
\begin{subfigure}[b]{0.45\textwidth}
    \centering
    \includegraphics[scale=0.65]{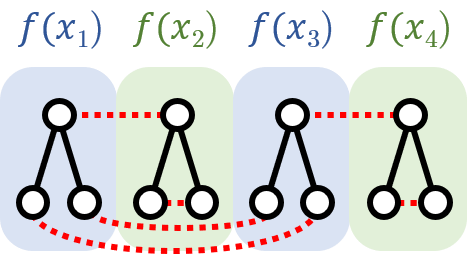}
    \caption{}
\end{subfigure}
\hfill
\caption{Forests representing two possible contractions of a set of weights to $C_{4,0}(x_1,x_2,x_3,x_4)=\lexpp{\theta}f(x_1) f(x_2) f(x_3) f(x_4)\rexp$ (see Figure~ \ref{fig:example_trees} for an explanation of this visual representation of a contraction). Consider the correlation function $C_{4,2}(x_1,x_2,x_3,x_4) = \sum_{\mu_1,\mu_2} \lexpp{\theta} \partial_{\mu_1} \partial_{\mu_2} f(x_1) \partial_{\mu_1} f(x_2) \partial_{\mu_2} f(x_3) f(x_4) \rexp$. \textbf{(a)} A contraction contributing to $C_{4,0}$ that does not also contribute to $C_{4,2}$ because it does not have a pairing corresponding corresponding the derivative contraction $\sum_{\mu_2} \partial_{\mu_2} f(x_1) \partial_{\mu_2}f(x_3)$, which requires at least one pairing between the weights belonging to $f(x_1)$ and $f(x_3)$. 
\textbf{(b)} A contraction contributing to $C_{4,0}$ that does contribute to $C_{4,2}$ because for each derivative contraction in $C_{4,2}$ their is at least one pairing between the weights of the corresponding $f(x)$.
\label{fig:forced_pairing}}
\end{figure}

Finding the set of contractions $P$ that contribute to a given correlation function is more non-trivial than the derivative-free case. 
However, as alluded to in the examples of Section \ref{sec:examples}, we claim \eqref{eq:derive_contractions} has the same asymptotic scaling as a \emph{subset} of contractions which contribute to an correlation function \emph{with no derivatives}.
Specifically,
\begin{lemma}
\label{lem:dt}
Consider a general correlation function $\tilde{C}(x_1,\dots,x_m) = \sum_{\rm indices} \lexpp{\theta} \prod_{K=1}^m F_K \rexp$, and let $C_0(x_1,\dots,x_m) = \lexpp{\theta} f_{\vec{p}_1,\Delta_1}(x_1) \cdots f_{\vec{p}_m,\Delta_m}(x_m) \rexp$ (the correlation function we get from $\tilde{C}$ if we drop all derivatives). The latter can be written as
\begin{align}
    C_0(x_1,\ldots,x_m) 
    % = \sum_{\mathrm{indices}} \lexpp{\theta} \prod_{K=1}^m f_{\vec{p}_K,\Delta_K}(x_K) \rexp 
    = \sum_{\vec{\cP}\in P_0} C_{\vec{\cP}}(x_1,\ldots,x_m) \,,
    \label{eq:lem2_1}
\end{align}
where $P_0$ is the set of contractions.
Then the correlation function $\tilde{C}$ can be written as
\begin{align}
    \tilde{C}(x_1,\ldots,x_m) 
    % = \sum_{\mathrm{indices}} \lexpp{\theta} \prod_{K=1}^m \partial^{k_K} f_{\vec{p}_K,\Delta_K} (x_K) \rexp 
    =  \sum_{\vec{\cP}\in P} a_{\vec{\cP}} C_{\vec{\cP}}(x_1,\ldots,x_m)\, .
    \label{eq:lem2_2}
\end{align}
Here, $a_{\vec{\cP}}$ are $n$-independent coefficients, and $P\subseteq P_0$ is a subset of contractions with the following property: If $F_I$ and $F_J$ have a pair of derivatives with shared indices, then every contraction in $P$ includes at least one pairing between weight factors of $f(x_I)$ and $f(x_J)$.

\end{lemma}
A proof of this lemma follows below.
See Figure~\ref{fig:forced_pairing} for examples of contractions from a $C_0$ that do and do not contribute to a $\tilde{C}$ with derivatives.

The strength of Lemma~\ref{lem:dt} lies in the fact that, since the $a_{\vec{\cP}}$ are $n$-independent constants, the asymptotic scaling of \eqref{eq:lem2_2} follows from calculating the asymptotic scaling of all contractions in the set $P$.
Finding the asymptotic scaling of all $\vec{\cP}\in P$ will then lead to the statement of Lemma~\ref{lem:main}.
Lemma \ref{lem:dt} is a generalization of the result for linear networks of \citet{dyer_asymptotics_2019}. 

\subsection{Proof of Lemma~\ref{lem:dt}}

Our technique for proving Lemma~\ref{lem:dt} will be to relate the set of contractions that contribute to an expectation with some derivative contraction to a \emph{subset} of the contractions that contribute to an expectation without said derivative contraction.

\begin{proof} [Proof (Lemma~\ref{lem:dt}).] 
Consider the behavior of an expectation with two derivatives with respect to $V_i$, where we sum over all possible $V$ weight derivatives,
\begin{align}
    \sum_{i=1}^{n}\mathbb{E}_{\theta}\left[\frac{\partial f_{\vec{p}_1,\Delta_1}}{\partial V_i}\frac{\partial f_{\vec{p}_2,\Delta_2}}{\partial V_i} \prod_{K=3}^m f_{\vec{p}_K,\Delta_K} \right]\, ,
    \label{eq:V_weight_derivs}
\end{align}
where without loss of generality we have assumed the derivatives act on $f_{\vec{p}_1,\Delta_1}$ and $f_{\vec{p}_2,\Delta_2}$.
One can evaluate this expression in a manner that is almost identical to \eqref{eq:ex_by_weight}.
That is, one can expand out all the distinct weights belonging to different layers and evaluate their individual expectations using Isserlis' Theorem. 
The only expectation that will be affected by the derivatives is that containing the $V$ weights.
Let us isolate that contribution and see how the derivatives affect its evaluation,
\begin{align}
    \sum_{i=1}^n \mathbb{E}_\theta \left[\frac{\partial V_{i^L_1}}{\partial V_i} \frac{\partial V_{i^L_2}}{\partial V_i} V_{i^L_3} \cdots V_{i^L_m} \right] &= \sum_{i=1}^n \lexpp{\theta} \delta_{i,i^L_1} \delta_{i,i^L_2} V_{i^L_3} \cdots V_{i^L_m}
    \rexp \, , \nonumber \\
    & = \delta_{i^L_1, i^L_2} \lexpp{\theta} V_{i^L_3} \cdots V_{i^L_m} \rexp \, , \nonumber \\
    & = \lexpp{\theta} V_{i^L_1} V_{i^L_2}  \rexp  \lexpp{\theta} V_{i^L_3} \cdots V_{i^L_m} \rexp \,. 
\end{align}
In the last line, we have rewritten the delta function as a pairing between the two $V$ weights the derivatives originally acted upon.
One is now  free to evaluate the expectation $\lexpp{\theta} V_{i^L_3} \cdots V_{i^L_m} \rexp$ using Isserlis's theorem.
Restoring all the other terms that were unaffected by the presence of derivatives, we thus find \eqref{eq:V_weight_derivs} is equal to a sum over a set of contractions that we call $P$.

Noticeably, every contraction $\vec{\cP}\in P$ will contain a pairing between $V_{i^L_1}$ and $V_{i^L_2}$, due to the factor $\lexpp{\theta} V_{i^L_1} V_{i^L_2} \rexp$.
This forced pairing between $V_{i^L_1}$ and $V_{i^L_2}$ is the net effect of the summed $V$ weight-derivatives.
Without the derivatives, we can obtain the same set of contractions $P$ by finding the \emph{subset} of all contractions contributing to the expectation $ \lexpp{\theta} \prod_{K=1}^m f_{\vec{p}_K,\Delta_K} \rexp$ in which each contraction has the $V$ weights belonging to $f_{\vec{p}_1,\Delta_1}$ and $f_{\vec{p}_2,\Delta_2}$ paired, i.e. all $C_{\vec{\cP}}$ for $\vec{\cP}$ in the subset contain a factor of $\lexpp{\theta} V_{i^L_1} V_{i^L_2} \rexp$.

It is easy to see that the above argument readily generalizes to correlation functions that include with respect to other weights.
For example, for two derivatives with respect to $W^{(\ell)}_{ij}$ weights, we can isolate the $W^{(\ell)}$ expectation in the same manner as we did above (now for arbitrary $f_{\vec{p}_I,\Delta_I}$ and $f_{\vec{p}_J,\Delta_J}$),
\begin{align}
    \sum_{i,j=1}^n &\mathbb{E}_\theta \left[\frac{\partial}{\partial W_{ij}}\left(W^{(\ell)}_{i^\ell_I j^{(\ell-1)}_{I,1}} \cdots W^{(\ell)}_{i^\ell_I j^{(\ell-1)}_{I,p_{I,\ell}}}\right) \frac{\partial}{\partial W_{ij}} \left( W^{(\ell)}_{i^\ell_J j^{(\ell-1)}_{J,1}} \cdots W^{(\ell)}_{i^\ell_J j^{(\ell-1)}_{J,p_{J,\ell}}}\right) \cdots W^{(\ell)}_{i^\ell_m j^{(\ell-1)}_{m,p_{m,\ell}}} \right] \, , \nonumber \\
    & = \delta_{i^\ell_I, i^\ell_J} \delta_{j^{(\ell-1)}_{I,1},j^{(\ell-1)}_{J,1}} \lexpp{\theta} \left(W^{(\ell)}_{i^\ell_I j^{(\ell-1)}_{I,2}}  \cdots W^{(\ell)}_{i^\ell_I j^{(\ell-1)}_{I,p_{I,\ell}}}\right)\left(W^{(\ell)}_{i^\ell_J j^{(\ell-1)}_{J,2}} \cdots W^{(\ell)}_{i^\ell_J j^{(\ell-1)}_{J,p_{J,\ell}}}\right) \cdots 
    \rexp + \cdots \, , \nonumber \\
    & = \lexpp{\theta} W^{(\ell)}_{i^\ell_I j^{(\ell-1)}_{I,1}} W^{(\ell)}_{i^\ell_J j^{(\ell-1)}_{J,1}}  \rexp  \lexpp{\theta} \left(W^{(\ell)}_{i^\ell_I j^{(\ell-1)}_{I,2}}  \cdots W^{(\ell)}_{i^\ell_I j^{(\ell-1)}_{I,p_{I,\ell}}}\right)\left(W^{(\ell)}_{i^\ell_J j^{(\ell-1)}_{J,2}} \cdots W^{(\ell)}_{i^\ell_J j^{(\ell-1)}_{J,p_{J,\ell}}}\right) \cdots 
    \rexp + \cdots \, ,
    \label{eq:lem2_derivs}
\end{align}
where the ``$+\cdots$'' in the second line onward represent the $p_{I,\ell}\times (p_{J,\ell} - 1)$ other terms that come from acting with the derivatives.
Despite the fact that this expression has significantly more terms, it is straightforward to see all terms in the sum of \eqref{eq:lem2_derivs} contain one factor of $\lexpp{\theta}W^{(\ell)}_{i^\ell_I j^{(\ell-1)}_{I,r}} W^{(\ell)}_{i^\ell_J j^{(\ell-1)}_{J,s}}\rexp$ for some $r=1,\ldots,p_{I,\ell}$ and $s=1,\ldots,p_{J,\ell}$.
We are again free to apply Isserlis' theorem at this point and recollect all the derivative-free terms, yielding an expression for the expectation written as a sum over a set of contractions.
However, it should be noted that one can obtain the same $\cP$ multiple times in \eqref{eq:lem2_derivs} when one applies Isserlis' theorem to all the $p_{I,\ell}\times p_{J,\ell}$ different derivative terms.
The potential degeneracy amounts to a coefficient, $a_{\vec{\cP}}$, in front of a given $C_{\vec{\cP}}$.
The degree of degeneracy is not important for our purposes, but crucially \emph{this coefficient has no $n$ dependence} since the number of redundant contractions is only dependent on $p_{I,\ell}$ and $p_{J,\ell}$.

Again, we can obtain the same set of contractions by finding a subset of all possible contractions that contribute to the same expectation without $W^{(\ell)}$ derivatives. 
Each $\vec{\cP}$ in this subset must contain at least one $W^{(\ell)}$ pairing between the weight belonging to $f_{\vec{p}_I,\Delta_I}$ and $f_{\vec{p}_I,\Delta_I}$, i.e. they contain a pairing of the form $\lexpp{\theta}W^{(\ell)}_{i^\ell_I j^{(\ell-1)}_{I,r}} W^{(\ell)}_{i^\ell_J j^{(\ell-1)}_{J,s}}\rexp$ for any $r=1,\ldots,p_{I,\ell}$ and $s=1,\ldots,p_{J,\ell}$. 
The same process above applies to $U$ weights as well. 

Thus, summing over all possible weight derivatives in the networks yields
\begin{align}
\sum_\mu \lexpp{\theta}\frac{\partial f_{\vec{p}_I,\Delta_I}}{\partial \theta^\mu}\frac{\partial f_{\vec{p}_J,\Delta_J}}{\partial \theta^\mu} \cdots \rexp = \sum_{\vec{\cP}\in P} a_{\vec{\cP}} C_{\vec{\cP}} (x_1,\ldots,x_m)\,,
\end{align}
where the $a_{\vec{\cP}}$ are $n$-independent coefficients and $P$ is the subset of the contractions that contribute to $\lexpp{\theta}\prod_{K=1}^m f_{\vec{p}_K,\Delta_K} \rexp$ where there is at least one pairing between the weights belonging to $f_{\vec{p}_I,\Delta_I}$ and $f_{\vec{p}_J,\Delta_J}$.

Generalizing this procedure to expectations that contain more than one derivative contraction is straightforward.
It is easy to see that each additional derivative pair will result in an additional pairing requirement on the set of contractions that contribute to the correlation function.

\end{proof}

\subsection{Proof of Lemma~\ref{lem:main}}

Using Lemma \ref{lem:dt}, calculating the asymptotic behavior of a correlation functions with derivatives is equivalent to calculating some subset of contractions, $P$, of the same correlation function without derivatives.
We will now show that all contractions in $P$ have contraction graphs with some minimum number of components, $N$.
Specifically, we will relate this minimum number of components to the properties of the correlation function's cluster graph, namely $N \leq n_e + n_0/2$.

\begin{proof} [Proof (Lemma~\ref{lem:main}).]
Consider a correlation function $C(x_1,\dots,x_m)$ with some number of derivative contractions. The cluster graph has $n_e$ even components, $n_o$ odd components, and edges $E$ determined by the derivative contractions (see Definition~\ref{def:cluster_graph}).

Using Lemma~\ref{lem:dt}, one can write this correlation function as a sum over a set of contractions, where the contractions are some subset of contractions that contribute to the same correlation function without derivatives.
Call the set of contractions that contribute to said correlation function $P$, so by Lemma \ref{lem:dt}
\begin{align}
    \tilde{C}(x_1,\dots,x_m) & = \sum_{\vec{\cP}\in P} a_{\vec{\cP}} C_{\vec{\cP}}(x_1,\dots,x_m) \, .
\end{align}

We now consider the properties of the contraction graphs corresponding to the contractions in $P$.
Let $\vec{\cP} \in P$.
The cluster graph of the correlation function is a subgraph of the contraction graph $\Gamma_{\vec{\cP}}$.
To see this, first note that each graph has $m$ vertices, in one-to-one correspondence with the factors $F_1,\dots,F_m$ appearing in the correlation functions.
The edges of the cluster graph have a one-to-one correspondence with the derivative pairs that appear in the correlation function.
And, Lemma~\ref{lem:dt} shows that every derivative pair in the correlation function is mapped to a pairing $\vec{\cP}$, which in turn is mapped to an edge in $\Gamma_{\vec{\cP}}$.
Therefore, the cluster graph is a subgraph of $\Gamma_{\vec{\cP}}$, and the number of components in $\Gamma_{\vec{\cP}}$ obeys $N \le n_e + n_o$.

Next, note that $\Gamma_{\vec{\cP}}$ only has components with even size.
This follows from the fact that each of $F_1,\dots,F_m$ has one $V$ weight.
The pairing of these weights in $\vec{\cP}$ is mapped to $m/2$ edges in $\Gamma_{\vec{\cP}}$ connecting the vertices in pairs.
This is a subset of all the edges in $\Gamma_{\vec{\cP}}$, and as a result all components have even size.
See for example Figure~\ref{fig:contraction_graph}.

What is the maximum number of components the contraction graph can have?
Each even size component in the cluster graph can be a component in the contraction graph, leading to $n_e$ components.
If we connect such components to other components, that would only reduce the number of components in the contraction graph.
On the other hand, an odd size components in the cluster graph cannot be a component of the contraction graph $\Gamma_{\vec{\cP}}$: it must be a non-trivial subgraph of some even sized component in $\Gamma_{\vec{\cP}}$.
For these odd components, the maximum number of contraction graph components is obtained if we connect them in pairs, ending up with $n_o/2$ components.
Therefore, the number of components in the contraction graph obeys $N \le n_e + n_o/2$.

\begin{figure}
\centering
\begin{subfigure}[b]{0.45\textwidth}
    \centering
    \includegraphics[scale=0.65]{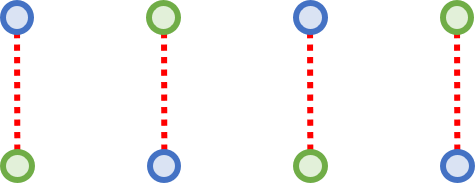}
    \caption{}
\end{subfigure}
\hfill
\begin{subfigure}[b]{0.45\textwidth}
    \centering
    \includegraphics[scale=0.65]{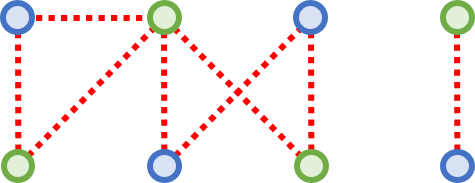}
    \caption{}
\end{subfigure}
\hfill
\caption{Example contraction graphs for $m=8$. \textbf{(a)} A contraction graph where we have \emph{only} shown edges associated with $V$-weight pairings. 
Since each vertex is associated with an $F_K$ with only a single $V$ weight factor, the $V$ pairings between different weight factor sets partition the contraction graph into components of size two. 
\textbf{(b)} Adding any additional weight pairings result in some number of components, all of which must have an even number of vertices. For the case shown, $N=2$.
\label{fig:contraction_graph}}
\end{figure}

\end{proof}

\section{Covariance asymptotics}
\label{app:Cov_asym}
In this appendix we state and discuss a theorem that allows one to bound the covariance of two products of derivative tensors. 
Next we state a related corollary that gives the variance of a product of derivative tensors. 
Lastly, we provide numerical results and give an explicit example of a variance calculation for the product of derivative tensors in $C_{4,2}$.

We can use Theorem \ref{conj:ethan guy conj fd} to bound the covariance of two sets of derivative tensors as well,
\begin{theorem}
\label{thm:cov}
Consider two correlation functions $C_x(x_{1},\ldots,x_{m_x})=\mathbb{E}_\theta[\mathcal{F}_x(x_1,\ldots,x_{m_{x}})]$ and $C_y(y_{1},\ldots,y_{m_y})=\mathbb{E}_\theta [\mathcal{F}_y(y_1,\ldots,y_{m_y})]$, with $\mathcal{F}_x$ and $\mathcal{F}_y$ some products of derivative tensors. 
For brevity, denote their arguments as $\tilde{x} := x_1,\ldots,x_{m_x}$ and $\tilde{y} := y_{1},\ldots,y_{m_y}$. 
Let the cluster diagrams of $C_x$ and $C_y$ have $(n_{e,x},n_{o,x})$ and $(n_{e,y},n_{o,y})$ even and odd components, respectively. 
If $n_{o,x}+n_{o,y}$ is even, then
\begin{align}
\emph{Cov}_{\theta}\left[\mathcal{F}_x(\tilde{x})\mathcal{F}_y( \tilde{y})\right]=\mathcal{O}(n^{s_V})\,,
\end{align}
where
\begin{align}
\label{eq:s_var}
s_V=\begin{cases}
s_{C,x}+s_{C,y}-1 & ,\, n_{o,x}+n_{o,y}=0 \\
s_{C,x}+s_{C,y} & ,\, n_{o,x}+n_{o,y}\ne 0
\end{cases}
\end{align}
where $s_{C,x}=n_{e,x}+\frac{1}{2}(n_{o,x}-m_{x})$ and $s_{C,y}=n_{e,y}+\frac{1}{2}(n_{o,y}-m_{y})$. 
If $n_{o,x}+n_{o,y}$ is odd, then $\emph{Cov}_{\theta}\left[\mathcal{F}_x(\tilde{x}) \mathcal{F}_y(\tilde{y})\right]=0$.
\end{theorem}

Note for the case where both correlation function are the same, we obtain the variance of $\mathcal{F}_x$.
Combined with Theorem \ref{thm:main}, this is useful in deducing the convergence properties of $\mathcal{F}_x$ in the large width limit; see below \citep{dyer_asymptotics_2019}.

\begin{proof}[Proof (Theorem~\ref{thm:cov}).]
It will be useful to write the covariance as
\begin{align}
    \text{Cov}_{\theta}\left[\mathcal{F}_x(\tilde{x}) \mathcal{F}_y(\tilde{y})\right] &= \mathcal{A}(\tilde{x}, \tilde{y})-\mathcal{B}(\tilde{x} , \tilde{y} ) \, ,
    \label{eq:covar}
\end{align}
where we have defined the various contributions to the covariance to be
\begin{subequations}
\begin{align}
\mathcal{A}(\tilde{x}, \tilde{y})	& := \lexpp{\theta}\mathcal{F}_x (\tilde{x}) \mathcal{F}_y (\tilde{y})\rexp \, , \\
\mathcal{B}(\tilde{x} , \tilde{y}) & := \lexpp{\theta} \mathcal{F}_x (\tilde{x}) \rexp \lexpp{\theta} \mathcal{F}_y ( \tilde{y}) \rexp \,.
\end{align}
\end{subequations}
Each of these correlation functions can be viewed as a sum over contractions of the form
\begin{subequations}
\begin{align}
    A_p(\tilde{x}, \tilde{y})	& := \sum_{\rm indices} \lexpp{\theta} \left(\prod_{K_x=1}^{m_x} F_{K_x} \right)\left(\prod_{K_y=1}^{m_y} F_{K_y} \right) \rexp  \, , \\
    B_p(\tilde{x}, \tilde{y}) & := \sum_{\rm indices} \lexpp{\theta} \prod_{K_x=1}^{m_x} F_{K_x} \rexp \lexpp{\theta} \prod_{K_y=1}^{m_y} F_{K_y} \rexp \, ,
\end{align}
\end{subequations}
where we are again using the shorthand $F_K := \partial^{k_K} f_{\vec{p}_K,\Delta_K}(x_K)$.
Each term is uniquely specified by the set of $\{ \{\vec{p}_{K_x},\Delta_{K_x}\}_{K_x = 1,\ldots,m_x}, \{\vec{p}_{K_y},\Delta_{K_y}\}_{K_y = 1,\ldots,m_y} \}$ as well as the derivative contraction structure of $\mathcal{F}_x$ and $\mathcal{F}_y$, which we have collectively denoted by a subscript $p$.
Then, to bound the asymptotic behavior of $\mathcal{A}$ and $\mathcal{B}$ it is enough to bound $A_p$ and $B_p$ for all $p$.

For a fixed $p$, note the only difference between $A_p$ and $B_p$ is that the former contains all weights in a single expectation, whereas the weights are split between two expectations in the latter.
Isserlis' Theorem can be applied to these correlation functions which will give several permutations of contractions,
\begin{align}
    A_p(\tilde{x}, \tilde{y})	& = \sum_{\vec{\cP}\in P^A_p} a_{\vec{\cP}} C_{\vec{\cP}}\left(\tilde{x}, \tilde{y}\right) \, , \\
    B_p(\tilde{x}, \tilde{y}) 
    % & = \left[\sum_{\vec{\cP}\in P^B_{x,p}} a_{\vec{\cP}} C_{\vec{\cP}}\left(\tilde{x}\right)\right]\left[\sum_{\vec{\cP}\in P^B_{y,p}} a_{\vec{\cP}} C_{\vec{\cP}}\left(\tilde{y}\right)\right] \, , \nonumber \\
    & = \sum_{\vec{\cP}\in P^B_p} a_{\vec{\cP}} C_{\vec{\cP}}\left(\tilde{x}, \tilde{y}\right) \, .
\end{align}
Due to the subtraction in \eqref{eq:covar}, any contractions which are present in both $P^A_p$ and $P^B_p$ (and have the same coefficient $a_{\vec{\cP}}$) will cancel out in the covariance.

Since $B_p$ contains two separate correlation functions, only \emph{disconnected} contractions, those which have no pairings between the weights contained in the $F_{K_x}$  and $F_{K_y}$, will contribute. 
Meanwhile, the contractions that contribute to $A_p$ will include these disconnected contractions but also others, since now the $F_{K_x}$ and $F_{K_y}$ are contained within a single correlation function and their weights can be paired.
Thus,  $P_p^B \subseteq P_p^A$. 
After their subtraction, the only contributions to the covariance that will not vanish are those contractions with at least one pairings between the weights belonging to a $F_{K_x}$  and a $F_{K_y}$, which we will call \emph{connected}.

From Lemma~\ref{lem:dt}, we know we can isolate contractions that have at least one pairing between certain derivative tensors by introducing derivatives with respect to weights.
Thus, to find the asymptotic scaling of only the connected pairings, we can consider the correlation function
\begin{align}
\label{eq:tildeC}
\tilde{C}_p \left(\tilde{x}, \tilde{y}\right) & := \sum_{\rm indices}\sum_{\mu}\mathbb{E}_{\theta}\left[\frac{\partial}{\partial\theta^{\mu}}\left(\prod_{K_x=1}^{m_x} F_{K_x} \right)\frac{\partial}{\partial\theta^{\mu}}\left(\prod_{K_y=1}^{m_y} F_{K_y} \right)\right] \,.
\end{align}
The derivative contraction in $\tilde{C}_p$ forces at least one pairing between the weights in the $F_{K_x}$ and those in the $F_{K_y}$, and thus all contractions that contribute to $\tilde{C}_p$ also contribute to $A_p - B_p$, up to $n$-independent coefficients. 
One can resum all the possible $p$, and so the asymptotic scaling of $\text{Cov}_{\theta}\left[\mathcal{F}_x(\tilde{x}) \mathcal{F}_y(\tilde{y})\right]$ is the same as that of
\begin{align}
    \sum_{\rm indices}\sum_{\mu} \lexpp{\theta} \frac{\partial \mathcal{F}_x \left(\tilde{x}\right)}{\partial \theta^\mu}\frac{\partial \mathcal{F}_y \left(\tilde{y}\right)}{\partial \theta^\mu} \rexp\, .
    \label{eq:cov_equiv}
\end{align}

Using Theorem~\ref{thm:main}, the asymptotic behavior of \eqref{eq:cov_equiv} can be found from the asymptotic behavior of $\mathcal{F}_x$ and $\mathcal{F}_y$. 
We consider two separate cases:
\begin{enumerate}
\item If the cluster graphs of $C_x$ and $C_y$ both have at least one odd component, then a connected graph can be constructed by connecting one odd component in $C_x$ to one odd component in $C_y$. 
Note that this does not change the scaling relative to the disconnected case, because it trades two odd components with one even component,
\begin{align}
(n_{e,x}+n_{e,y},n_{o,x}+n_{o,y})\to(n_{e,x}+n_{e,y}+1,n_{o,x}+n_{o,y}-2)\, .
\end{align}
\item Instead, if all components in the cluster graphs of $C_x$ and $C_y$ are even, then to form a connected component we must choose connect an even component in $C_x$ to an even component in $C_y$, resulting in one fewer even components,
\begin{align}
(n_{e,x}+n_{e,y},0 )\to(n_{e,x}+n_{e,y}-1,0) \,.
\end{align}
\end{enumerate}
These results give the scalings on the right-hand side of \eqref{eq:s_var}.
\end{proof}

For the case where $\mathcal{F}_x= \mathcal{F}_y$ Theorem \ref{thm:main} reduces to a bound on the variance of a given correlation function,
\begin{cor}
\label{cor:var}
Define a correlation function $C(x_1,\ldots,x_m)=\mathbb{E}_\theta[\mathcal{F}(x_1,\ldots,x_m)]$ with $\mathcal{F}$ some product of derivative tensors. If the cluster graph of $C$ has $(n_e,n_o)$ even and odd components the variance of $\mathcal{F}$ is
\begin{align}
\emph{Var}_{\theta}\left[\mathcal{F}(x_1,\ldots,x_m)\right]=\mathcal{O}(n^{s_V})\, ,\qquad s_V:=\begin{cases}
2s_{C}-1 & n_{o}=0\, ,\\
2s_{C} & n_{o}\ne 0\, .
\end{cases}
\label{eq:variance}
\end{align}
where $s_C=n_e + (n_o-m)/2$ is the correlation function's scaling exponent, $C=\mathcal{O}(n^{s_C})$.
\end{cor}
For $n_{o}\ne0$, this result is in agreement with \citet{dyer_asymptotics_2019}, but for cases where $n_o=0$, this result is a tighter bound.

Numerical results of the variance of various products of derivative tensors are shown in Table \ref{tab:numerics_var}.
Note these agree with the tighter bound of Corollary \ref{cor:var}.

\begin{table}
\centering
\begin{tabular}{c|cc|ccc}
  $C$ & $(n_e,n_o)$ & predicted exponent & tanh & sigmoid & softplus \\ %& ELU\\
\hline
$C_{2,0}$ & $(0,2)$ & $0$  & $-0.01$ & $0.02$ & $-0.09$ \\ %& $-0.04$ \\
$C_{2,1}$ & $(1,0)$ & $-1$  & $-0.995$ & $-1.01$ & $-0.99$ \\ %& $-0.02$ \\
$C_{4,0}$ & $(0,4)$ & $0$  & $0.02$ & $0.06$ & $-0.05$ \\ %& $0.30$ \\
$C_{4,2}$ & $(0,2)$ & $-2$  & $-1.998$ & $-1.99$ & $-2.07$ \\ %& $-0.98$ \\
$C_{4,3}$ & $(1,0)$ & $-3$  & $-3.05$ & $-3.01$ & $-3.08$ \\ %& $-1.00$ \\
$C_{6,4}$ & $(0,2)$ & $-4$  & $-4.00$ & $-4.01$ & $-4.01$ \\ %& $-2.03$ 
\end{tabular}
\medskip
\caption{Numerical results for variance of the given correlation functions. 
The experimental setup is the same as that of Table~\ref{tab:smooth}.
The predicted exponent is $s_V$ in \eqref{eq:variance}. \label{tab:numerics_var}}

\end{table}

\paragraph*{Case Study $C_{4,2}$:}

\begin{figure}
\centering
\begin{subfigure}[b]{0.33\textwidth}
    \centering
    \includegraphics[scale=0.45]{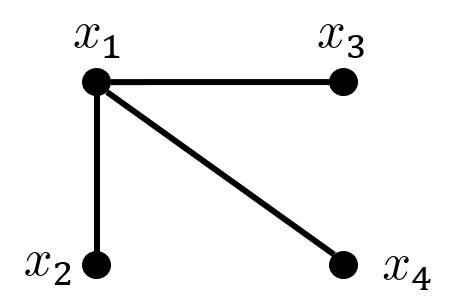}
    \caption{}
\end{subfigure}
\hfill
\begin{subfigure}[b]{0.66\textwidth}
    \centering
    \includegraphics[scale=0.45]{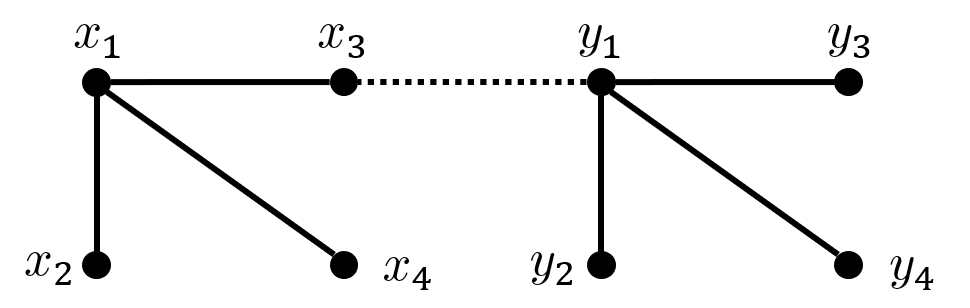}
    \caption{}
\end{subfigure}
\hfill
\caption{Cluster graphs for \textbf{(a)} $C_{4,2} = \mathbb{E}_\theta[\mathcal{F}_{4,2}]$ and \textbf{(b)} $\text{Var}_\theta(\mathcal{F}_{4,2})$. 
The dotted black line represents the derivative contraction introduced through \eqref{eq:tildeC}. 
This line could run from any vertex in $\{x_1,x_2,x_3,x_4 \}$ to any vertex in $\{y_1,y_2,y_3,y_4 \}$. 
However, no matter how one chooses such a connection, this will result in a single component. 
This gives an overall sclaing of $\mathcal{O}(n^{-3})$ instead of the naive doubling of $s_C$ which yields $\mathcal{O}(n^{-2})$. 
\label{fig:covariance}}
\end{figure}

For $C_{4,2} = \lexpp{\theta}\mathcal{F}_{4,2}\rexp$, defined in \eqref{eq:C42}, we have $(n_{e},n_{o})=(1,0)$ and from Corollary \ref{cor:var}, this falls in the case where we should observe an extra $-1$ in the scaling of its variance. 
The additional derivative to enforce contractions between the disconnected parts means the scaling is equivalent to that of a correlation function with a single component of size $8$, see Figure~\ref{fig:covariance}. 
Using Theorem~\ref{thm:main}, the scaling is
\begin{align}
n_{e}+\frac{n_{o}}{2}-\frac{m}{2}=1-\frac{8}{2}=-3\,.
\end{align}
Note this agrees with numerics (see Table \ref{tab:numerics_var}) and provides a tigher bound than Lemma 5 of \citet{dyer_asymptotics_2019}, which would have predicted $\text{Var}_\theta [\mathcal{F}_{4,2}]=\mathcal{O}(n^{-2})$. 

\section{Piecewise-linear activations}
\label{app:pwl}

To understand the discrepancy in the scaling for the linear case (see Table \ref{tab:smooth}), we consider the behavior of correlation functions with higher-rank derivative tensors. 
Consider once more the example network of Section~\ref{sec:examples} (with $L=1$ and $d=1)$. 
An example of a rank-two derivative tensors is
\begin{align}
\frac{\partial^{2}f(x)}{\partial U_i \partial U_j} & = \frac{2}{\sqrt{n}}\sum_{k=1}^n \delta_{ik}\delta_{jk} V_k x^2\, .
\label{eq:double u derive fd}
\end{align}
Contrast this result with the linear case, where the right-hand side of \eqref{eq:double u derive fd} would vanish because $\frac{\partial f(x)}{\partial U_i} = \frac{1}{\sqrt{n}} V_i x$ has no more $U$-weight dependence. 
This is important for particular correlation functions that contain contributions from terms like \eqref{eq:double u derive fd}. 
Let us take a look at a case where such a term is relevant.

\paragraph*{Case Study $C_{4,3}$:}

Consider once more the correlation function $C_{4,3}$, defined in \eqref{eq:C43}. 
This is a correlation function where there appears to be some numerical difference in the asymptotic behavior depending on the activation function.
In particular, comparing  the results in Table~\ref{tab:smooth}, we see the function appears to have different scaling for the real analytic versus the piecewise-linear activation functions: $\mathcal{O}(n^{-1})$ for the former and $\mathcal{O}(n^{-2})$ for the latter. 

We consider an $L=2$ network with $d=1$. 
In the linear case, since repeated application of the same derivative causes a derivative tensor to vanish, the only non-vanishing rank-three derivative tensor is $\frac{\partial^3 f(x)}{\partial V_i \partial W_{jk} U_{l}} = \frac{1}{n} \delta_{ij} \delta_{kl} x$. 
A straightforward calculation, described below, shows that the scaling of said correlation function is $\mathcal{O}(n^{-2})$.
 
We can see if this changes for an $L=2$ non-linear network, namely the case where $\phi(x)=x^2$. 
Since repeated weight derivatives do not automatically vanish, many more terms can contribute to $C_{4,3}$. 
The asymptotic behavior of an example term in the correlation function can also be calculated explicitly, 
\begin{align}
\sum_{i,j,k=1}^n & \lexpp{\theta}\frac{\partial^{3}f(x_1)}{\partial U_i \partial U_j \partial U_k}\frac{\partial f(x_2)}{\partial U_i}\frac{\partial f(x_3)}{\partial U_j}\frac{\partial f(x_4)}{\partial U_k}\rexp =\cO(n^{-1}) \,.
\label{eq:C3 one hidden fd}
\end{align}
This is evidence that the difference in asymptotic behavior observed for certain correlation functions in Table~\ref{tab:smooth} results from terms with repeated weight derivatives.
For piecewise linear activations, such contributions vanish.

%Since we can confirm the numerical discrepancy of $n$-scaling analytically, it appears a tighter bound might be possible for activation functions which vanish after a finite number of derivatives. 
%Numerics seem to indicate this also includes the ReLU and other piecewise linear activations. 
%We leave such an analysis for future work.

We now describe the calculation of $C_{4,3}$ in detail.

\paragraph*{Linear Case.}

We consider $C_{4,3}$ for the two-hidden layer linear network (i.e. $\phi(x)=x$) with $d=1$. 
The network functional is $f_\theta(x) = \frac{1}{n}\sum_{i,j}V_i W_{ij} U_j x$.
The only non-vanishing rank-three derivative tensor for this network is given by
\begin{align}
\frac{\partial^3 f(x)}{\partial V_i \partial W_{jk} \partial U_l} & = \frac{1}{n} \delta_{ij} \delta_{kl} x\,.
\end{align}
As such, the only non-vanishing contribution to $C_{4,3}$ is given by
\begin{align}
C_{4,3}(x_1,x_2,x_3,x_4) & = \sum_{i,j,k,l=1}^n \lexpp{\theta}\frac{\partial^{3}f(x_1)}{\partial V_{i}\partial W_{jk}\partial U_l}\frac{\partial f(x_2)}{\partial V_{i}}\frac{\partial f(x_3)}{\partial W_{jk}}\frac{\partial f(x_4)}{\partial U_l}\rexp \, , \nonumber \\
&= \frac{1}{n^4}\sum_{i,j,k,l,j_2,i_4}^{n}\lexpp{\theta} \delta_{ij} \delta_{kl} W_{ij_2}U_{j_2}  V_j U_k V_{i_4}W_{i_4l}\rexp x_1 x_2 x_3 x_4 \, , \nonumber \\ 
&= \frac{1}{n^2} x_1 x_2 x_3 x_4 = \mathcal{O}(n^{-2}) \,.
\end{align}
Note that since the cluster graph of $C_{4,3}$ has $(n_e,n_0)=(1,0)$, its predicted scaling is $\cO(n^{-1})$.
Hence, this result is not in violation of the predicted asymptotic behavior, but this is an example where the bound is not tight.

\paragraph*{Polynomial Case.}

Let us now calculate $C_{4,3}$ for the same network but with activations $\phi(x)=x^2$. 
The network functional is now
\begin{align}
    f_\theta(x) & =\frac{1}{\sqrt{n}}\sum_{i=1}^n V_i \left(\frac{1}{\sqrt{n}}\sum_{j=1}^n W_{ij}\left(U_j x\right)^2\right)^2 \, , \nonumber \\
    & =\frac{1}{n^{3/2}}\sum_{i,j,k=1}^n V_i W_{ij} W_{ik} U_j^2 U_k^2 x^4 \,.
\end{align}
Since repeated derivatives do not automatically vanish, there are many more contributions.
For example, there are derivatives such as
\begin{align}
% \frac{\partial f(x)}{\partial U_i} & = \frac{2}{n^{3/2}} \sum_{l,p,q =1}^n V_l W_{lp} W_{lq} \left(\delta_{ip}U_p U_q^2 + \delta_{iq}U_q U_p^2\right)x^4 \, , \\
\frac{\partial^2 f(x)}{\partial U_i \partial U_j} & = \frac{2}{n^{3/2}} \sum_{l,p,q =1}^n V_l W_{lp} W_{lq} \left(\delta_{ip}\delta_{jp}U_q^2+4\delta_{iq}\delta_{jp}U_p U_q + \delta_{iq}\delta_{jq}U_p^2\right)x^4 \, , \\
\frac{\partial^3 f(x)}{\partial U_i \partial U_j \partial U_k} & = \frac{4}{n^{3/2}} \sum_{l,p,q =1}^n V_l W_{lp} W_{lq} \nonumber \\
& \quad\times \left(\delta_{ip}\delta_{jp}\delta_{kq} U_q+2\delta_{iq}\delta_{jp}\delta_{kp} U_q + 2\delta_{iq}\delta_{jp}\delta_{kq} U_p + \delta_{iq}\delta_{jq}\delta_{kp}U_p\right)x^4 \, ,
\end{align}
We will consider one particular contribution
that has different scaling than was found in the linear case. 
Explicitly, we have
\begin{align}
C_{4,3} & \supset \sum_{i,j,k}\lexpp{\theta}\frac{\partial^{3}f_\theta(x_1)}{\partial U_i \partial U_j \partial U_k}\frac{\partial f_\theta(x_2)}{\partial U_i}\frac{\partial f_\theta(x_3)}{\partial U_j}\frac{\partial f_\theta(x_4)}{\partial U_k}\rexp\nonumber \\
& = \frac{32}{n^6} \sum_{i,j,k,\vec{i},\vec{j},\vec{k}}^n \lexpp{\theta}V_{i_1}V_{i_2}V_{i_3}V_{i_4}\rexp \nonumber \\
& \quad \times \lexpp{\theta}W_{i_1j_1}W_{i_1k_1}W_{i_2j_2}W_{i_2k_2}W_{i_3j_3}W_{i_3k_3}W_{i_4j_4}W_{i_4k_4}\rexp \nonumber \\
& \quad \times \mathbb{E}_{\theta}\Big[\left(\delta_{ij_1}\delta_{jj_1}\delta_{kk_1} U_{k_1}+2\delta_{ik_1}\delta_{jj_1}\delta_{kj_1} U_{k_1} + 2\delta_{ik_1}\delta_{jj_1}\delta_{kk_1} U_{j_1} + \delta_{ik_1}\delta_{jk_1}\delta_{kj_1}U_{j_1}\right) \nonumber \\
& \quad \times \left(\delta_{ij_2}U_{j_2} U_{k_2}^2 + \delta_{ik_2}U_{k_2} U_{j_2}^2\right)\left(\delta_{jj_3}U_{j_3} U_{k_3}^2 + \delta_{jk_3}U_{k_3} U_{j_3}^2\right)\left(\delta_{kj_4}U_{j_4} U_{k_4}^2 + \delta_{kk_4}U_{k_4} U_{j_4}^2\right) \Big] \nonumber \\
& \quad \times x_1^4 x_2^4 x_3^4 x_4^4\, ,
\end{align}
where $\vec{i}$ is shorthand for the indices $i_1,i_2,i_3,i_4$, and similarly for $\vec{j}$ and $\vec{k}$.
Obviously, this expressions is a mess.
However, with some foresight, we pick out a particular term from the various contractions that come from applying Isserlis' theorem,
\begin{align}
C_{4,3} & \supset \frac{64}{n^6} \sum_{i,j,k,\vec{i},\vec{j},\vec{k}}^n \lexpp{\theta}V_{i_1}V_{i_2}\rexp\lexpp{\theta}V_{i_3}V_{i_4}\rexp \nonumber \\
& \quad \times \lexpp{\theta}W_{i_1k_1}W_{i_2j_2}\rexp\lexpp{\theta}W_{i_1j_1}W_{i_2k_2}\rexp\lexpp{\theta}W_{i_3j_3}W_{i_4j_4}\rexp\lexpp{\theta}W_{i_3k_3}W_{i_4k_4}\rexp \nonumber \\
& \quad \times \delta_{ik_1}\delta_{jj_1}\delta_{kj_1}\delta_{ij_2}\delta_{jj_3} \delta_{kj_4} \lexpp{\theta} U_{k_1} U_{j_2} \rexp\lexpp{\theta}U_{k_2}^2\rexp\lexpp{\theta}U_{j_3} U_{j_4} \rexp\lexpp{\theta}U_{k_3}^2 \rexp\lexpp{\theta}U_{k_4}^2\rexp \nonumber \\
& \quad \times x_1^4 x_2^4 x_3^4 x_4^4 \,, \nonumber \\
& = \frac{64}{n^6} \sum_{\vec{i},\vec{j},\vec{k}}^n \delta_{i_1i_2}\delta_{i_3i_4} \times \delta_{k_1j_2}\delta_{j_1k_2}\delta_{j_3j_4} \delta_{k_3k_4}  \times \delta_{k_1j_2}\delta_{j_1j_3}\delta_{j_1j_4}\delta_{k_1j_2} \delta_{j_3j_4} \times x_1^4 x_2^4 x_3^4 x_4^4 \, , \nonumber \\
& = \frac{64}{n} x_1^4 x_2^4 x_3^4 x_4^4\, .
\end{align}
Thus we see this is $\cO(n^{-1})$, and hence $C_{4,3}$ is $\cO(n^{-1})$ for this network functional.

\end{appendix}

\end{document}